\documentclass[a4paper,fleqn]{cas-dc}

\usepackage[numbers]{natbib}
\usepackage{subcaption}
\usepackage[vlined, ruled, boxed, linesnumbered]{algorithm2e}

\newcommand{\veryshortarrow}[1][3pt]{\mathrel{%
   \vcenter{\hbox{\rule[-.5\fontdimen8\textfont3]{#1}{\fontdimen8\textfont3}}}%
   \mkern-4mu\hbox{\usefont{U}{lasy}{m}{n}\symbol{41}}}}

\newcommand{\MYhref}[3][blue]{\href{#2}{\color{#1}{#3}}}%

\def\tsc#1{\csdef{#1}{\textsc{\lowercase{#1}}\xspace}}
\tsc{WGM}
\tsc{QE}
\tsc{EP}
\tsc{PMS}
\tsc{BEC}
\tsc{DE}

\begin{document}
\let\WriteBookmarks\relax
\def\floatpagepagefraction{1}
\def\textpagefraction{.001}
\shorttitle{Discriminative Feature Alignment}
\shortauthors{Jing Wang et~al.}

\title [mode = title]{Discriminative Feature Alignment: Improving Transferability of Unsupervised Domain Adaptation by Gaussian-guided Latent Alignment}                      

\author[1]{Jing Wang}[orcid=0000-0001-9417-1174]
\cormark[1]
\ead{jing@ece.ubc.ca}


\address[1]{Department of Electrical and Computer Engineering, University of British Columbia , Vancouver, BC, Canada}

\author[2]{Jiahong Chen}[orcid=0000-0001-7152-8230]
\ead{jhchen@mech.ubc.ca}

\author[1]{Jianzhe Lin}
\ead{jianzhelin@ece.ubc.ca}

\address[2]{Department of Mechanical Engineering, University of British Columbia , Vancouver, BC, Canada}

\author[3]{Leonid Sigal}
\ead{lsigal@cs.ubc.ca}

\address[3]{Department of Computer Science, University of British Columbia , Vancouver, BC, Canada}

\author[2]{Clarence W. de~Silva}
\ead{desilva@mech.ubc.ca}

\cortext[cor1]{Code is available at \MYhref{https://github.com/JingWang18/Discriminative-Feature-Alignment}{https://github.com/JingWang18/Discriminative-Feature-Alignment}}

\begin{abstract}
In this paper, we focus on the unsupervised domain adaptation problem where an approximate inference model is to be learned from a labeled data domain and expected to generalize well to an unlabeled data domain. The success of unsupervised domain adaptation largely relies on the cross-domain feature alignment. Previous work has attempted to directly align latent features by the classifier-induced discrepancies. Nevertheless, a common feature space cannot always be learned via this direct feature alignment especially when a large domain gap exists. To solve this problem, we introduce a Gaussian-guided latent alignment approach to align the latent feature distributions of the two domains under the guidance of the prior distribution. In such an indirect way, the distributions over the samples from the two domains will be constructed on a common feature space, i.e., the space of the prior, which promotes better feature alignment. To effectively align the target latent distribution with this prior distribution, we also propose a novel unpaired L1-distance by taking advantage of the formulation of the encoder-decoder. The extensive evaluations on nine benchmark datasets validate the superior knowledge transferability through outperforming state-of-the-art methods and the versatility of the proposed method by improving the existing work significantly.
\end{abstract}


\maketitle

\section{Introduction}
The performance of computer vision models has been improved significantly by deep neural networks that take advantage of large quantities of labeled data. However, the models trained on one dataset typically perform poorly on another, different, but related, dataset \cite{unbiased, pan2009survey}. This shortcoming calls for adaptation strategies that help transfer knowledge from a label-rich source domain to a label-scarce target domain. Among such adaptation strategies, unsupervised domain adaptation (UDA) aims at mitigating domain shift in a way that does not use the target dataset labels, while attempting to maximize the performance of the classifier on them. Existing UDA algorithms attempt to mitigate domain shifts by only considering the classifier-induced discrepancy between the two domains, which can reduce the domain divergence \cite{ben-david01}. Both adversarial \cite{dsn,dann,cycle,saito,adda} and non-adversarial domain adaptation (DA) \cite{mmdlong,norm} methods work under the guidance of convergence learning bounds \cite{ben-david01}. The main idea behind these bounds is that concurrently minimizing the source domain classification error and the classifier-induced discrepancy between the source domain and the target domain, inadvertently aligning the two latent feature spaces in which classification is done. In particular, adversarial DA attempts to align the feature spaces by minimizing the classifier-induced discrepancy with adversarial objectives. 

\begin{figure}[t]
\begin{center}
\includegraphics[width=1.0\linewidth]{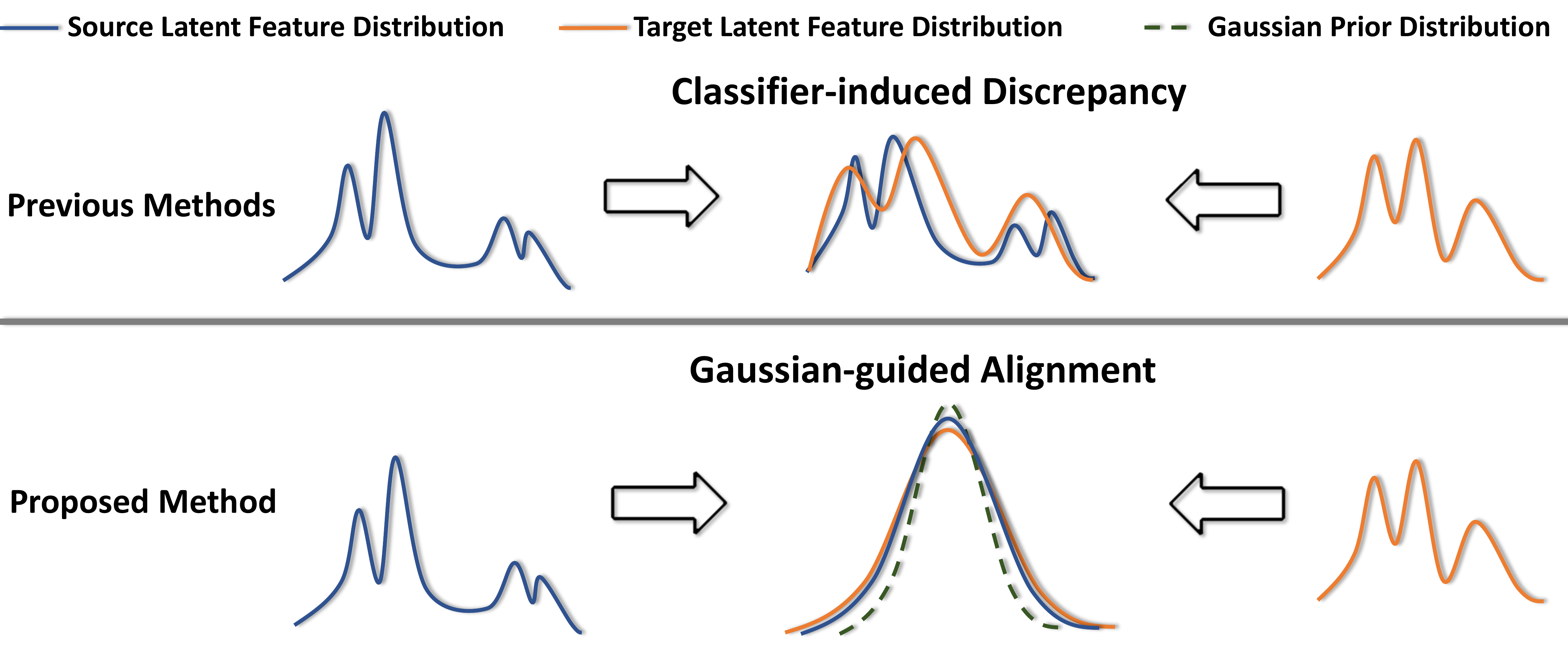}
\end{center}
   \caption{\textbf{(Best viewed in color.)} Existing UDA methods try to align the feature distributions of the two domains by the classifier-induced discrepancies. However, it might be difficult for them to construct the two feature distributions in a single distribution space or align arbitrarily complex feature distributions in that space. Our method attempts to {\em indirectly} align the features of the two domains under the guidance of the Gaussian prior distribution. Our method can encourage the features of the two domains to be constructed in a common feature space, i.e., the space of the Gaussian prior, where the target samples can maximally take advantage of the discriminative source features for their own classification tasks.}
\label{fig:distribution}
\end{figure}

However, as shown in Figure~\ref{fig:distribution}, adaptation in this manner alone cannot effectively learn a common feature space for the classification in the two domains. This claim is empirically validated in Section~\ref{sec:ab_shape}. To address this problem, we propose a \emph{discriminative feature alignment} (DFA) to align the two latent feature distributions of the source dataset and the target dataset under the guidance of the Gaussian prior (similar to VAE \cite{vae}). Because the classification takes place in the latent space, the latent space itself is discriminative, in turn, making alignment focus on the discriminative feature distributions. Our approach is built on the encoder-decoder (autoencoder) formulation with an implicitly shared discriminative latent space (see Figure~\ref{fig:framework}). Specifically, we define a feature extractor $G$ which takes and encodes input samples into a latent space; similarly we define a decoder $D$ which takes a latent feature vector, or a random vector sampled from a Gaussian prior, and decodes it back to the image. Both the encoder ($G$) and the decoder ($D$) are shared by the samples from the source domain and the target domain; and one can consider $D$ as a form of regularization. We utilize a KL-divergence penalty to encourage the latent distribution over the source samples to be close to the Gaussian prior. While we can similarly encourage the target distribution in the feature space to be close to the Gaussian prior, thereby achieving the desired alignment, this turns out less effective in practice. Instead, the alignment between the source and target distributions in the latent space is achieved by a novel unpaired L1-distance between the reconstructed samples from the decoder, i.e., minimizing the distance between $D(G(\mathbf{x}_s))$ and $D(G(\mathbf{x}_t))$ among all pairs of samples from the source domain (s) and the target domain (t). The proposed regularization for the distribution alignment is named \emph{distribution alignment loss}. We further find that instead of aligning the latent distributions directly, we get better results by aligning the target latent distribution to the Gaussian prior, i.e., minimizing the distance between $D(G(\mathbf{x}_t))$ and the decoded samples from the prior in the feature space. The sampling also serves as data augmentation and could be useful in scenarios where the source dataset itself maybe limited. 

Moreover, the proposed DFA can be incorporated into other UDA frameworks, either adversarial or non-adversarial, to improve results via a better feature alignment. To validate the versatility of DFA, we demonstrate it using an adversarial framework for the digit classification and a non-adversarial framework for the object classification. The two frameworks are developed based on the existing techniques, mainly: maximum classifier discrepancy (MCD) \cite{saito} and stepwise adaptive feature norm (SAFN) \cite{norm}, since they are state-of-the-art for the digit classification and the object classification, respectively. In all settings, our DFA significantly improves the performance of the original frameworks and outperforms other existing frameworks by a large margin.

\vspace{0.1in}
\noindent
{\bf Contributions:}
\begin{itemize}
\item We propose a novel model for unsupervised domain adaptation, which utilizes an {\em indirect} latent alignment process to construct a common feature space under the guidance of a Gaussian prior. 
\item We introduce a new method to align two distributions, which, 
instead of minimizing discriminator error using a GAN,
minimizes the direct L1-distance between the decoded samples. 
\item We evaluate the proposed frameworks and the versatility of the proposed DFA on both digit and object classification tasks by adapting it into existing UDA approaches, and achieve state-of-the-art performance on the benchmark datasets.
\end{itemize}

\section{Related Work}
Existing UDA methods can be divided into two major types: adversarial and non-adversarial domain adaptation. 

\subsection{Adversarial Domain Adaptation}
Motivated by generative adversarial nets (GANs) \cite{gan}, adversarial DA methods, which stem from the technique proposed in \cite{dann}, are widely explored by the DA community. The goal is for the latent feature distributions of the two domains to be aligned, such that domain classifier is unable to recognize domain from which the features originate. 
In early works, such alignment was realized by simple batch normalization statistics, which aligned the data distributions from the two domains to a canonical form \cite{Autodial2017F,Revisiting2016Li}. 
Introducing an adversarial loss 
makes it more difficult for the domain classifier to classify the domains correctly \cite{gta}, producing better alignment. Further advances in adversarial DA can be found in recent works. Long et al. propose to measure the domain divergence by considering the distribution correlations for each class of objects \cite{Partial2018Long,CADA2018Long,Multi2018Long}. Domain separation network \cite{dsn} is also proposed to better preserve the component that is private to each domain before aligning the latent feature distributions.

However, the mechanism concerns constructing adversarial learning between the feature extractor and the domain classifier, which does not consider the relationship between the decision boundary and the target samples. Maximum classifier discrepancy (MCD), instead, involves an adversarial mechanism between its image classifiers and the feature extractor \cite{saito}. This method can align the latent feature distributions of the two domains by considering the decision divergence on predicting the target samples between the two image classifiers.

\subsection{Non-adversarial Domain Adaptation}

Existing non-adversarial DA methods attempt to quantify domain shifts by designing specific statistical distances between the two domains. Correlation alignment \cite{coral,deepcoral} utilizes the difference of the mean and the covariance between the two datasets as the domain divergence, and attempts to match them during the training. The methods based on maximum mean discrepancy (MMD) \cite{mmd} such as \cite{mmdlong,jointlong} measure the variance between the latent feature distributions of the two domains. Some studies \cite{chen2019progressive,saito2017asymmetric,zhang2018collaborative} also propose to learn the discriminative representations by pseudo-labels and aligning the output class distributions. However, they still consider classifier-induced discrepancies for the latent alignment, which cannot guarantee the safe transfer of the discriminative features across domains. Moreover, stepwise adaptive feature norm (SAFN) \cite{norm} identifies that domain shifts rely on the less-informative features with small norms for the target-specific task, and the knowledge across domains can be safely transferred by placing the target features far away from these small-norm regions.  

\section{Method}
\label{sec:method}
In this section, the details of the proposed method are presented. First, we discuss the preliminary of the UDA problem in Section~\ref{sec:pre}. Second, we explain about the way to achieve knowledge transfer by taking advantage of the formulation of the encoder-decoder in Section~\ref{sec:decoder}. Third, we discuss the overall idea of the proposed model in Section~\ref{sec:framework}. Fourth, we give details about the loss functions that are used in the proposed method in Section~\ref{sec:loss}. Finally, we demonstrate the versatility of the proposed method by incorporating it into the existing UDA methods.
\subsection{Preliminary}\label{sec:pre}
Under the setting of UDA, we sample $n$ labeled images from the source space $\{X_S, Y_S\}$ to form the source domain $\mathfrak{D}_S=\{(\mathbf{x}^{(i)}_{\mathbf{s}}, \mathbf{y}^{(i)}_{\mathbf{s}})\}_{i=1}^n$, as well as $m$ unlabeled images from the target space $\{X_T, Y_T\}$ to form the target domain $\mathfrak{D}_T=\{(\mathbf{x}^{(j)}_{\mathbf{t}})\}_{j=1}^m$. The objective of UDA is to obtain a feature extractor $G$ that generates a target distribution in the feature space that can maximize the performance of classifying $\mathbf{x_t}$ without accessing its label.

\subsection{Knowledge Transfer via Encoder-Decoder}\label{sec:decoder}
\begin{figure}[h]
    \center
    \includegraphics[width=1.0\linewidth]{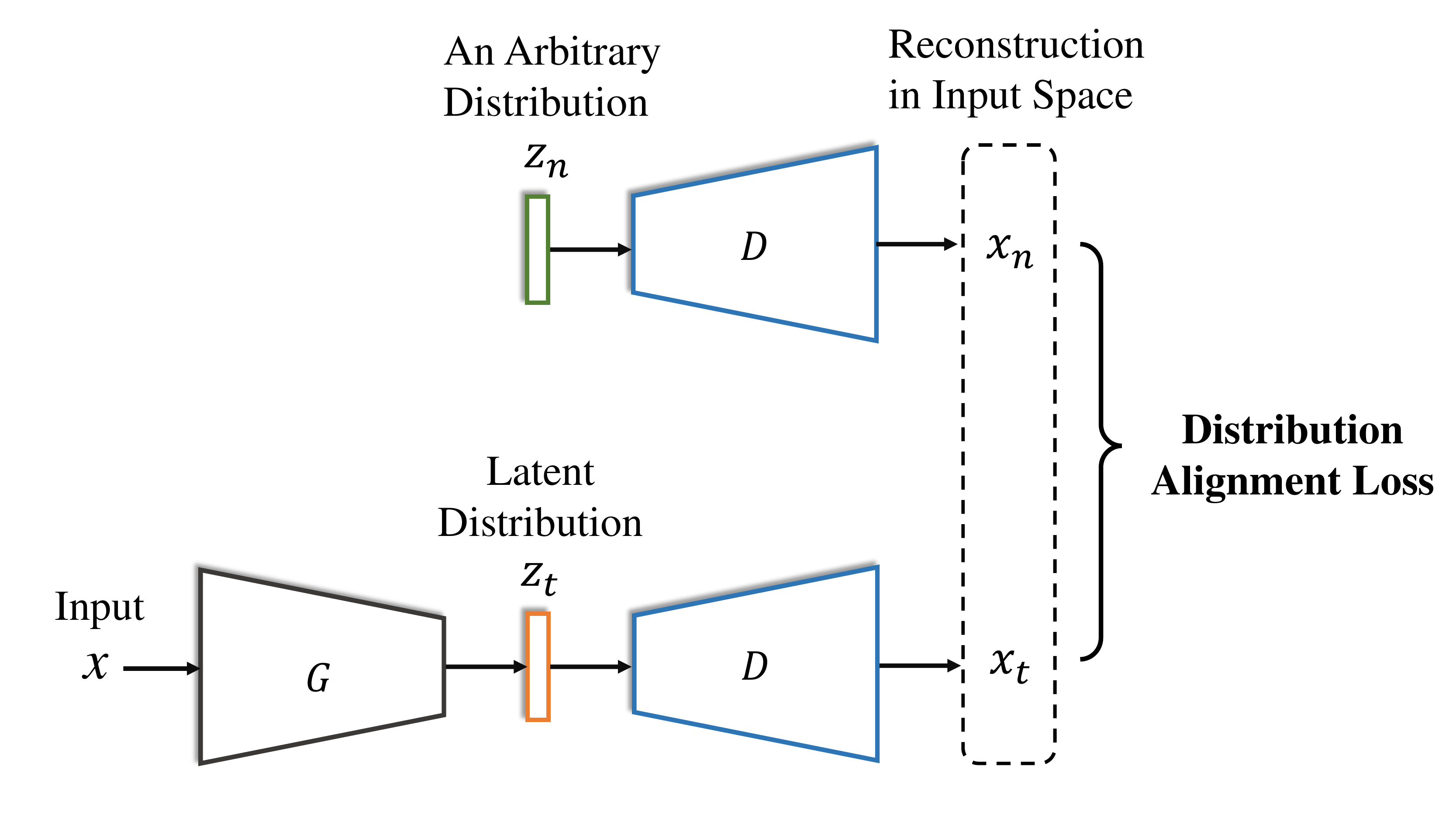}
    \caption{Aligning two distributions by taking advantage of the formulation of the encoder-decoder. It contains an encoding function $G$ and a decoding function $D$. The mapping function $D(G(\circ))$ can be regarded as a weight-tied autoencoder that can put the less representative features into the nonlinear regime of $G$'s nonlinearity.}\label{fig:idea}
\end{figure}

The proposed work is under the assumption that every neural-network-based UDA framework should consist of a feature extractor $G$ and an image classifier $F$. The goal of the proposed method is not only to align the latent distributions of the two domains but also to make $G$ learn the representation from the target samples under the guidance of the discriminative source representation. As illustrated in Figure~\ref{fig:idea}, the decoder $D$ is specifically used for the proposed \emph{distribution alignment loss} to align the target latent distribution with the prior distribution. Thus, $G$ is also an encoder that learns the hidden representations for both $F$ and $D$ in our setting. As $G$ continuously shares its learning parameters with $D$ during the training, our model can also be viewed as a weight-tied autoencoder. The proposed \emph{distribution alignment loss}, which is different from the reconstruction loss used in the existing work on autoencoder, is an L1-distance between the reconstructed target samples and the decoded samples from the prior in the feature space.

\subsubsection{Knowledge Transfer via Distribution Alignment}
The objective of unsupervised domain adaptation is to retain sufficient knowledge about the source domain in the target latent space. In a single-domain problem, the information about the input domain can be retained in its latent space by reconstructing the input samples \cite{denoiseae}. Motivated by this, we argue that minimizing the difference between the reconstructed target samples and the source input samples can encourage the target latent space to cover sufficient information about the source domain. To be specific, minimizing the proposed \emph{distribution alignment loss} on the premise of constructing the source feature space on the space of the prior is equivalent to maximizing the lower bound of the mutual information between the latent space of the target domain and the input space of the source domain .

In the setting of UDA, we are interested in learning the correspondence between the samples from the target latent space $Z_T$ and the samples from the source input space $X_S$:
\begin{equation}\label{eqn:encoding-decoding}
\begin{split}
&X_T\xrightarrow{G_\theta}Z_T\xrightarrow{D_\theta}\hat{X}_{T}\\
&X_S\xrightarrow{G_\theta}Z_S\xrightarrow{D_\theta}\hat{X}_{S},
\end{split}
\end{equation}
where the encoder $G$ shares it learning parameters $\theta$ with the decoder $D$. 

The mutual information between the source input space and the target latent space can be expressed as
\begin{equation}\label{eqn:mi-source-target}
\mathsf{I}(X_S;Z_T) = \mathsf{H}(X_S) -  \mathsf{H}(X_S|Z_T),
\end{equation}
where $\mathsf{I}(\cdot)$ is the mutual information; $\mathsf{H}(\cdot)$ is the entropy. $\mathsf{H}(X_S)$ is an unknown constant since the source input space $X_S$ is from a fixed distribution that will not be affected by $\theta$. Hence, the information maximization process can be reduced according to Equation~\ref{eqn:mi-source-target}:
\begin{equation}
\begin{split}
\max_\theta\mathsf{I}(X_S;Z_T) &= \max_\theta -  \mathsf{H}(X_S|Z_T)\\
&=\max_\theta\mathbb{E}_{p(X_S,Z_T)}[\log p(X_S|Z_T;\theta)].
\end{split}
\end{equation}

Normally, the reconstructed target sample $\hat{x}_{t} = D_\theta(z_{t})$ is not exactly the same as a corresponding source sample $x_s$. However, in probabilistic terms, the parameters of a distribution $p(x_s|z_{t})$ may produce $\hat{x}_s$ with high probability as they share the same object feature. Therefore, the lower bound of the mutual information can be maximized by minimizing
\begin{equation}\label{eqn:domian-cons-loss-prob_1}
L_{1}(x_{s},\hat{x}_{t}) \propto -\log p(x_{s}|z_{t}),
\end{equation}
where $L_1$ is the L1 distance.

However, this objective cannot be achieved because of the lack of the correspondence between the reconstructed samples from the target domain and the input samples from the source domain. 

To tackle this problem, we define a prior distribution $q(\mathbf{z_n})$ and construct the discriminative source features on the space of the prior $Z_N$. If there exists $D_{KL}(q(\mathbf{z_n})||p(\mathbf{z_s})) = 0$, $Z_S \approx Z_N$, Equation~\ref{eqn:encoding-decoding} becomes 
\begin{equation}
\begin{split}\label{eqn:encoding-decoding-xn}
&X_T\xrightarrow{G_\theta}Z_T\xrightarrow{D_\theta}\hat{X}_{T}\\
&X_S\xrightarrow{G_\theta}Z_S\approx Z_N \xrightarrow{D_\theta}\hat{X}_{N}\approx\hat{X}_{S},
\end{split}
\end{equation}

Now, we define a distribution $q(\hat{X}_S|Z_T)$ for the following inequality:
\begin{equation}\label{eqn:lower-bound}
\mathbb{E}_{p(X_S,Z_T)}[\log p(X_S|Z_T)] \geq \mathbb{E}_{q(\hat{X}_S,Z_T)}[\log q(\hat{X}_S|Z_T)],
\end{equation}
where $\mathsf{D}_{KL}(q||p)\geq 0$.

The left-hand side of Equation~\ref{eqn:lower-bound} is the lower bound of the mutual information between the source input space and the target latent space. We thus have a new lower bound for the mutual information:
\begin{equation}\label{eqn:lower-bound-nonparametric}
    \max_\theta\mathsf{I}(X_S;Z_T) \geq \max_\theta\mathbb{E}_{q(\hat{X}_S,Z_T)}[\log q(\hat{X}_S|Z_T;\theta)].
\end{equation}
Considering the parametric distribution $q(\hat{X}_S|Z_T; \theta)$, the lower bound shown in Equation~\ref{eqn:lower-bound-nonparametric} can be maximized by
\begin{equation}\label{eqn:lower-bound-maximum}
\begin{split}
\max_{\theta} \mathbb{E}_{q(\hat{X}_S,Z_T)}[\log q(\hat{X}_S|Z_T; \theta)].
\end{split}
\end{equation}

Therefore, the mutual information $\mathsf{I}(X_S;Z_T)$ can be maximized when $\exists \theta$ s.t. $q(\hat{X}_S|Z_T;\theta) = p(X_S|Z_T;\theta)$. 

Combining Equation~\ref{eqn:encoding-decoding-xn} and Equation~\ref{eqn:lower-bound-maximum}, we have the lower bound of the mutual information between $X_S$ and $Z_T$ as maximizing
\begin{equation}\label{eqn:lower-bound-reconstruct}
\begin{split}
    &\mathbb{E}_{q(Z_N, X_T)}[\log q(\hat{X}_S \approx \hat{X}_N = D_{\theta}(Z_N)|Z_T=G_\theta(X_T))].
\end{split}
\end{equation}

Then, we consider the \emph{distribution alignment error}:
\begin{equation}\label{eqn:domian-cons-loss-prob}
L_{1}(\hat{x}_{n},\hat{x}_{t}) \approx L_{1}(\hat{x}_{s},\hat{x}_{t}) \propto -\log q(\hat{x}_{s}|z_{t}),
\end{equation}

We thus have the following minimization that is equivalent to the maximization of the lower bound of the mutual information:
\begin{equation}
\begin{split}
&\min_{\theta} \mathbb{E}_{q(\hat{X}_S, \hat{X}_T)}[L_1(\hat{X}_{S},\hat{X}_{T})]\\
\Rightarrow &\min_{\theta} \mathbb{E}_{q(Z_N, X_T)}[L_1(D_{\theta}(Z_N),D_\theta(G_\theta(X_T)))],
\end{split}
\end{equation}
which can be rewritten according to Equation~\ref{eqn:domian-cons-loss-prob_1} and Equation~\ref{eqn:domian-cons-loss-prob}:
\begin{equation}\label{eqn:max-min-distribution-aligment-error}
\begin{split}
&\max_\theta\mathsf{I}(X_S;Z_T) \\
\geq &\max_{\theta} \mathbb{E}_{q(\hat{X}_S, Z_T)}[\log q(\hat{X}_S|Z_{T};\theta)] \\
\approx&\max_{\theta} \mathbb{E}_{q(\hat{X}_N, Z_T)}[\log q(\hat{X}_N|Z_{T};\theta)] \\
=&\max_{\theta} \mathbb{E}_{q(Z_N, X_T)}[\log q(D_{\theta}(Z_N)|G_\theta(X_T))] \\
= &\min_{\theta} \mathbb{E}_{q(Z_N, X_T)}[L_1(D_{\theta}(Z_N),D_\theta(G_\theta(X_T)))] \\
\end{split}
\end{equation}

At this point, we can conclude that the lower bound of the mutual information between the source input space $X_S$ and the target latent space $Z_T$ can be maximized by minimizing the proposed distribution alignment error $L_{1}(\hat{x}_{n},\hat{x}_{t})$ on the premise that the source latent distribution is close enough to the prior.

\subsubsection{Decoder}
The proposed regularization has two functionalities in our model: 1) distribution alignment; 2) discriminative feature extraction. The distribution alignment mechanism alone cannot guarantee the produced latent distribution $p(\mathbf{z_t})$ is adequately discriminative for $F$ to generalize well to the target domain. To further enforce $G$ to focus on the cross-domain classification discriminative characteristics of the target samples, we let the weight matrices of $G$ and $D$ be symmetric. The choice of weight tying for the proposed encoder-decoder is motivated by the denoising autoencoder (DAE) \cite{denoiseae}. DAE shows that the tying weight makes it more difficult for an encoder to stay in the linear regime of its nonlinearity. 

We denote a mapping layer of $G$ followed by a nonlinearity $\sigma_{i}$ by
\begin{equation}
\begin{aligned}
g_{\theta}(\textbf{x}) = \sigma_{i}(\textbf{W}_{i}\textbf{x} + \textbf{b}_{i})
\end{aligned}
\end{equation}
with learning parameters $\theta = (\textbf{W}_{i},\textbf{b}_{i})$, where $\textbf{W}_{i}$ is the weight matrix for the convolutional layer and $\textbf{b}_{i}$ is its bias matrix.
Similarly, we define a mapping layer of $D$ followed by the same nonlinearity $\sigma_{i}$ as
\begin{equation}
\begin{aligned}
d_{\theta^{T}}(\textbf{y}) = \sigma_{i}(\textbf{W}_{i}^{T}\textbf{y} + \textbf{b}_{i}^{T})
\end{aligned}
\end{equation}
with learning parameters $\theta^{T} = (\textbf{W}_{i}^{T},\textbf{b}_{i}^{T})$, where $\textbf{W}_{i}^{T}$ is the weight matrix for the 2-D transposed convolutional layer and $\textbf{b}_{i}^{T}$ is its bias matrix. Therefore, without considering the pooling, unpooling and batch normalization, our $2L$-layer autoencoder with tying weight can be denoted by
\begin{equation}
\begin{aligned}
\hat{\textbf{x}} = \sigma_{1}(\textbf{W}_{1}^{T}(\dots\sigma_{L}(\textbf{W}_{L}^{T}(\sigma_{L}(\textbf{W}_{L}(\dots \sigma_{1}(\textbf{W}_{1}\textbf{x} + \textbf{b}_{1}) \\ + \dots) + \textbf{b}_{L}) + \textbf{b}_{L}^{T}) + \dots) + \textbf{b}_{1}^{T}),
\end{aligned}
\end{equation}

Then, with the support of a task-specific classifier, the less representative features can be placed in the nonlinear regime of the encoder $G$ and, therefore, rejected. As our objective is to encourage $p(\mathbf{z_t})$ to be as discriminative as possible, it is straightforward to take advantage of this property of weight tying. The layers with different functionalities of the proposed decoder $D$ are listed below: 

 \textbf{2-D Transposed Convolution} A convolutional layer can be represented as a sparse matrix $\textbf{W}$, and has $\textbf{W}^{T}$ for its backward propagation. Thus for $D$, we have a transposed convolutional layer $\textbf{W}^{T}$ that utilizes $\textbf{W}^{T}$ and $\textbf{W}$ for its forward and backward propagations, respectively. 
    
\textbf{Max Unpooling} The max unpooling used for $D$ takes the output, i.e., the maximum value, of the corresponding max pooling of $G$ and the indices of this output as its input. Then, the output of the max unpooling is appropriately sized by setting all non-maximal values to zero. While this type of operation is not a good inverse of the max pooling, it is perfectly suitable for our objective. This is because we only want to retain the features extracted by $G$ for the proposed \emph{distribution alignment loss}.

\textbf{Average Unpooling} The average unpooling utilized for $D$ takes the output of the corresponding average pooling of $G$ as its input and sets other values to this average. Similar to the max unpooling, this operation only maintains the information of the features extracted by $G$.

\textbf{Nonlinearity} We observed from our experiments that the nonlinearity term retained a significant amount of features that were extracted by $G$. Therefore, we assume that the impact of the nonlinearity is limited to the reconstruction of the hidden representation extracted from the target domain to achieve the distribution alignment. In this study, we use the same activation function for $D$ as that of $G$, i.e., ReLU activation, without considering the reversibility of the proposed encoder-decoder.

The average unpooling utilized for the decoder is the upsampling using the nearest-neighbor interpolation. The max unpooling used for the decoder is \emph{torch.nn.MaxUnpool2d}\footnote{https://pytorch.org/docs/stable/nn.html} implemented by \textbf{Pytorch}. The transposed convolution utilized for the decoder is \emph{torch.nn.functional.conv\_transpose2d}\footnote{https://pytorch.org/docs/stable/nn.functional} implemented by \textbf{Pytorch}. The tying weight is achieved by sharing the weight matrix of the corresponding convolution with the transposed convolution. Our decoder for the object classification tasks can be viewed as an inverted version of the feature extractor of ResNet-50 with 2-D transposed convolution and upsampling. The detailed architecture and configuration of the proposed ResNet-50-based decoder are presented in the Appendix.

\subsection{Framework of Discriminative Feature Alignment}\label{sec:framework}
In this section, we will discuss how to construct the latent distributions of the two domains on the space of the prior using the proposed regularization. 

\begin{figure*}
\begin{center}
\includegraphics[width=1.0\linewidth]{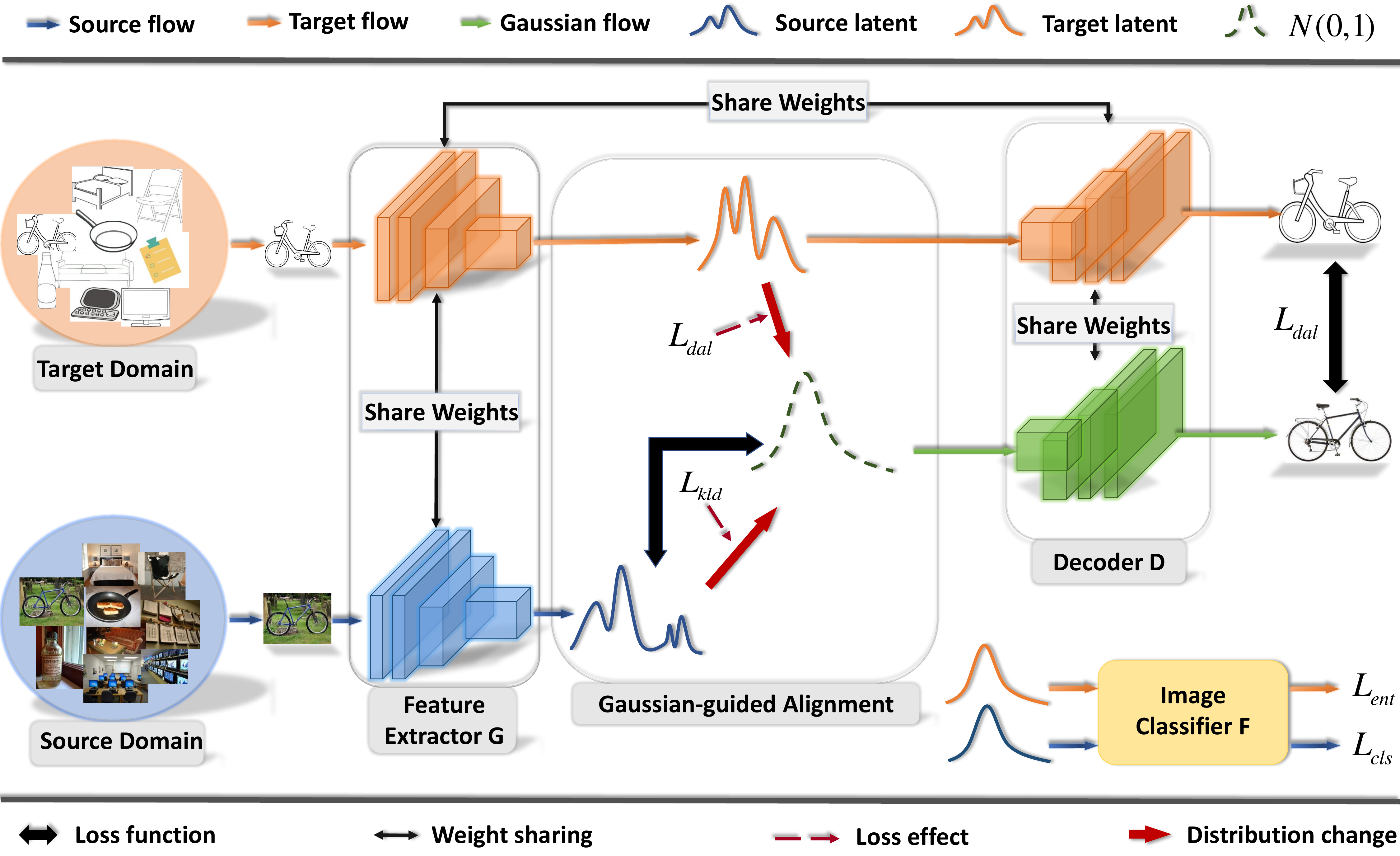}
\end{center}
   \caption{\textbf{(Best viewed in color.)} The overall architecture of the proposed framework. The feature extractor G maps the input data to their latent feature vectors. The decoder D, which can be viewed as an inverted version of $G$, maps a latent feature vector or Gaussian prior vector to an image that has the same dimensions as the input samples. Our model can encourage the discriminative features of the two domains to be projected into the space of the prior.}
\label{fig:framework}
\end{figure*}

Our model, as illustrated in Figure~\ref{fig:framework}, consists of a feature extractor $G$ and a decoder $D$ that share the learning parameters $\theta_g$. To predict the categories of the input samples, the framework developed based on our model should also have an image classifier $F$. We represent a mapping function from the input data, either $\mathbf{x_s}$ or $\mathbf{x_t}$, to its latent feature vector $\mathbf{z_s}$ or $\mathbf{z_t}$ as $G(\mathbf{x}; \theta_g)$. Meanwhile, we denote a mapping function from a latent feature vector or the Gaussian prior vector to an image by $D(\mathbf{z}; \theta_{g})$.

As the source dataset labels are accessible, we can make a reasonable assumption that the feature space of the source domain is discriminative. Therefore, the goal of our model is to learn a latent feature distribution $p(\mathbf{z_t})$ from the target domain that can maximally take advantage of the discriminative features of the source domain for its own classification. To achieve this, we need to design a feature alignment approach that can ultimately construct the two feature spaces in a common distribution space. The problem is how to define such distribution space and effectively project the features of the two domains into this space. 

For this objective, we propose to indirectly align the source features and the target features under the guidance of the Gaussian prior. As the first step of our model, we define the Gaussian prior distribution $q(\mathbf{z_n})\sim \mathcal{N}(0,1)$ where we will construct the two feature spaces on. To encourage the discriminative feature space of the source domain to be constructed on the space of the prior, we regularize $G$ and $F$ by softmax cross-entropy loss on the labeled source samples, and enforce the distribution over the source samples $p(\mathbf{z_s})$ to be close to the Gaussian prior $q(\mathbf{z_n})$ via the KL-divergence penalty on $G$. Meanwhile, the latent feature distribution of the target domain $p(\mathbf{z_t})$ should be similarly close to the Gaussian prior. In preliminary experiments, we tried to use the same KL-divergence penalty to achieve such alignment, but it turned out to be not as effective as we expected. Therefore, to effectively align $p(\mathbf{z_t})$ with the prior distribution $q(\mathbf{z_n})$, we propose a novel L1-distance between the reconstructed samples from the decoder, i.e., minimizing the distance between $D(G(\mathbf{x_t}))$ and $D(\mathbf{z_n})$, to regularize $G$. Once the training of our model converges, the three distributions, i.e., the source and the target distributions in the feature space and the Gaussian prior distribution, can be properly aligned. In other words, our method can effectively construct the feature spaces of the two domains in the same distribution space, i.e., the space of the Gaussian prior. We also include different ways to achieve such latent-space alignment in Section~\ref{sec:ablation} and compare them with our proposed method.

\subsection{Loss Functions}\label{sec:loss}
\subsubsection{Softmax Cross-entropy Loss}
We use softmax cross-entropy loss to handle the classification task on the labeled source domain. This objective can ensure that the discriminative feature space of the source domain can be properly constructed on the space of the prior. We train both $G$ and $F$ to minimize the objective function:
\begin{equation}
\begin{aligned}
\mathcal{L}_{cls}(X_{S}, Y_{S}) =  -\frac{1}{M} \sum_{i=1}^{M} I(i = \mathbf{y_s}^{(i)}) \log{p_{s}(\mathbf{x_s}^{(i)})},
\end{aligned}
\end{equation}
where $I(i = \mathbf{y_s}^{(i)})$ is a binary indicator which is 1 when $i$ equals $\mathbf{y_s}^{(i)}$; $p_{s}$ is the mapping function for the classification scores, i.e., $p_{s} = \mathbf{softmax} \circ F \circ G$.

\subsubsection{Kullback-Leibler Divergence}
To encourage the latent feature distribution of the source domain to be close to the Gaussian prior, we apply the KL-divergence penalty between $p(\mathbf{z_s})$ and $q(\mathbf{z_n})$ to regularize $G$. We express this objective as:
\begin{equation}
\begin{aligned}
\mathcal{L}_{kld}(X_{S}) =  \frac{1}{M} \sum_{i=1}^{M}q(\mathbf{z_n}^{(i)})\log{\frac{q(\mathbf{z_n}^{(i)})}{G(\mathbf{x_s}^{(i)})}},
\end{aligned}
\end{equation}
where G seeks to generate the discriminative features of the source domain in the space of the prior under the support of $\mathcal{L}_{cls}$.
\subsubsection{Distribution Alignment Loss}\label{sec:dal}
Regularizing $G$ and $F$ by $\mathcal{L}_{cls}$ and $\mathcal{L}_{kld}$, respectively, makes the discriminative feature space of the source domain be constructed on the space of the prior. Therefore, by encouraging $p(\mathbf{z_t})$ to be defined in the same distribution space, tasks on the target domain can maximally take advantage of the knowledge learned from the source labels. To achieve this, we propose a simple yet effective method to align the target latent distribution with the prior distribution, namely, \emph{distribution alignment loss} (DAL). DAL is applied to regularize both $G$ and $D$. We utilize the absolute difference between the two data distributions produced by $D$ and formulate the proposed DAL as:
\begin{equation}\label{eqn:domain-consistency-loss}
\begin{aligned}
\mathcal{L}_{dal}(X_T) = \frac{1}{M} \sum_{i=1}^{M} \lvert\lvert D(G(\mathbf{x_t}^{(i)});\theta_g) - D(\mathbf{z_n}^{(i)};\theta_g)\lvert\lvert_{1},
\end{aligned}
\end{equation}
where $\lvert\lvert \circ \lvert\lvert_{1}$ is the L1-norm. In Section~\ref{sec:syn}, we present a detailed analysis of the proposed DAL, and empirically verify that it serves as a distribution alignment mechanism. 

\subsubsection{Entropy Loss}
In the proposed framework DFA-ENT, the latent feature vector $\mathbf{z_t}$ is fed into $F$ to produce predictions for the target input samples. To control the contribution of the target predictions in the generalization of an image classifier, we employ a low-density separation technique \emph{entropy minimization} (ENT) \cite{ent} to measure the class overlap of the target samples:
\begin{equation}
    \mathcal{L}_{ent}(X_{T}) = \frac{1}{M}\sum_{i=1}^M -F(G(\mathbf{x_t}^{(i)}))\log F(G(\mathbf{x_t}^{(i)})).
\end{equation}

\subsubsection{Full Objective}
The full objective function of the proposed framework DFA-ENT is a linear combination of softmax cross-entropy loss, KL-divergence penalty, \emph{distribution alignment loss} and the entropy loss: 
\begin{equation}
    \mathcal{L} = \mathcal{L}_{cls} + \mathcal{L}_{ent} + \alpha \mathcal{L}_{kld} + \beta \mathcal{L}_{dal},
\end{equation}
where $\alpha$ and $\beta$ are the weights for the KL-divergence penalty and DAL,respectively, to control the relative importance of the proposed regularization. 

\subsection{Versatility}

\subsubsection{Adversarial Domain Adaptation}
\label{sec:ada} 
Maximum classifier discrepancy \cite{saito} achieves state-of-the-art on digit and traffic-sign classification. It has one feature extractor $G$ and two image classifiers $F_1$ and $F_2$. It regards the disagreement between $F_1$ and $F_2$ as its classifier-induced discrepancy. It uses a three-step adversarial training strategy to avoid the input target samples that are outside the support of the source domain: first, minimizing softmax cross-entropy loss $\mathcal{L}_{cls}$; second, minimizing the difference between $\mathcal{L}_{cls}$ and the L1-loss between the outputs of the two image classifiers on the target samples $\mathcal{L}_{adv}(X_T)$; and third, minimizing $\mathcal{L}_{adv}(X_T)$.

The proposed DFA-MCD is developed based on MCD. Our objective $\mathcal{L}_{kld}$ is integrated into the first and the second training steps of MCD; and the proposed $\mathcal{L}_{dal}$ is combined with the objective function of its last training step. To better clarify DFA-MCD, we include the details of the training procedures in Algorithm~\ref{alg:mcd} and highlight our method in red. 

\begin{algorithm}
\SetAlgoLined
\scriptsize
 Input image normalization; {\textcolor{red}{initialize the Gaussian prior $q(\mathbf{z_n}) \sim \mathcal{N}(0,1)$}}\;
 \While{epoch $\leq$ max epoch}{
  \For{$batch\gets1$ \KwTo N}{
   \textbf{Step 1}: {\textcolor{red}{Sample minibatch of $M$ samples from the Gaussian prior $q(\mathbf{z_n})$}}\; Update $G$, $F_1$ and $F_2$ to $\smash{\displaystyle\min_{G,F_1,F_2}[\mathcal{L}_{cls}(X_{S}, Y_{S})}${\textcolor{red}{$+\alpha\mathcal{L}_{kld}(X_{S})$}}]\;
   \texttt{\\}
   \textbf{Step 2}: Fix $G$; and update $F_1$ and $F_2$ to 
   $\smash{\displaystyle\min_{F_1,F_2}[\mathcal{L}_{cls}(X_{S}, Y_{S})} -\mathcal{L}_{adv}(X_{T})${\textcolor{red}{$+\alpha\mathcal{L}_{kld}(X_{S})$}}] \;
   \texttt{\\}
   \textbf{Step 3}: Fix $F_1$ and $F_2$. {\textcolor{red}{Calculate $\mathcal{L}_{dal}(X_T)$ using the current $\theta_g$}}. Then update $G$ {\textcolor{red}{and $D$}} to $\smash{\displaystyle\min_{G,D} [\mathcal{L}_{adv}(X_{T})}${\textcolor{red}{$+\beta\mathcal{L}_{dal}(X_{T})]$}}.
   }
 }
 \caption{DFA-MCD}
\label{alg:mcd} 
\end{algorithm}

\subsubsection{Non-adversarial Domain Adaptation}
Stepwise adaptive feature norm \cite{norm} is state-of-the-art approach on non-adversarial DA and object classification. It follows the standard DA setting with a feature extractor $G$ and a $l$-layer image classifier $F$. It denotes the first $l-1$ layers of its image classifier as $F_f$, and utilizes the intermediate features from $F_f$ to calculate its classifier-induced discrepancy:
\begin{equation}
L_{d}(x_{i}) = L_{2}(h(x_{i};\theta_{p}) + \delta{r}, h(x_{i};\theta_{c})),
\end{equation}
where $L_{2}$ is the L2-distance; $h(x)$ is the L2-norm of $F_{f}(G(x))$; 
$\theta_{p}$ and $\theta_{c}$ represent the learning parameters in the previous and the current iterations, respectively; and $\delta{r}$ is a constant to control the feature-norm enlargement. Thus, SAFN can mitigate domain shifts by minimizing the following loss:
\begin{equation}
\begin{split}
&\mathcal{L}_{safn}(X_S,Y_S,X_T) \\
=& \mathcal{L}_{cls}(X_S, Y_S) + \mathcal{L}_{ent}(X_T) + \kappa\mathbb{E}_{x_{i} \in (X_S \cup X_T)} [L_{d}(x_{i})],
\end{split}
\end{equation}
where $\kappa$ is a trade-off among the objectives.

Our DFA-SAFN is developed based on SAFN. We implement a ResNet-50-based decoder to generate $D(\mathbf{z_t})$ and $D(\mathbf{z_n})$ for the proposed DAL. We integrate all of our objective functions into the final loss of SAFN. The details of DFA-SAFN are shown in Algorithm~\ref{alg:safn}.

\begin{algorithm}
\SetAlgoLined
\scriptsize
 Input image normalization; initialize tensors for storing $h(x_{i};\theta_{p})$, {\textcolor{red}{$D(\mathbf{z_t})$ and $D(\mathbf{z_n})$; initialize the Gaussian prior $q(\mathbf{z_n}) \sim \mathcal{N}(0,1)$}}\;
 \While{epoch $\leq$ max epoch}{
  \For{$batch\gets1$ \KwTo N}{
    {\textcolor{red}{Sample minibatch of $M$ samples from the Gaussian prior $q(\mathbf{z_n})$}}\;
    Calculate $L_{d}(X_{S} \cup X_{T})$ using $h(x_{i};\theta_{p})$ and $h(x_{i};\theta_{c})$\;
    {\textcolor{red}{Calculate $\mathcal{L}_{dal}$ using $D(\mathbf{z_t})$ and $D(\mathbf{z_n})$ from the previous iteration}}\;
    Update $G$, $D$ and $F$ to minimize [$\mathcal{L}_{safn}+${\textcolor{red} {$\alpha\mathcal{L}_{kld}+\beta\mathcal{L}_{dal}$}}]\;
    Calculate $h(x_{i};\theta_{c})$ and store it as $h(x_{i};\theta_{p})$ for the next iteration\;
    {\textcolor{red}{Get $D(\mathbf{z_t})$ and $D(\mathbf{z_n})$ using the current $\theta_g$ for the next iteration}}\;
   }
 }
 \caption{DFA-SAFN}
\label{alg:safn}
\end{algorithm}

\section{Experiments}
We implemented all experiments on the \textbf{PyTorch}\footnote{https://pytorch.org/} platform. We reported the results of the benchmark algorithms under their optimal hyper-parameter settings. To better validate the versatility of our model, we followed the same settings and the hyper-parameters that were utilized in MCD \cite{saito} and SAFN \cite{norm} for evaluating DFA-MCD and DFA-SAFN, and did not fine-tune the two frameworks. To be specific, we used Adam \cite{kingma2014adam} optimizer, and set the learning rate and the batch size to $2.0 \times 10^{-4}$ and 128, respectively, in all experiments for the evaluation on the digit and traffic-sign recognition datasets; we utilized SGD optimizer, and set the learning rate and the batch size to $1.0 \times 10^{-3}$ and 32, respectively, in all experiments for the evaluation on the object recoginition benchmark datasets.

\subsection{Experiments on Synthetic Datasets}\label{sec:syn}
In this section, we empirically verified the distribution alignment mechanism of the proposed \emph{distribution alignment loss} (DAL) on three synthetic datasets, namely, 2D Gaussian distributions with different mean or covariance, \emph{moons} dataset and \emph{blobs} dataset. For each experiment, we generated 500 samples for each domain. We employed the same networks $G$ and $D$ for all synthetic experiments. The encoder $G$ is a 3-layer MLP that maps a 2D distribution to a higher dimensional space. The deocder $D$, which is also a 3-layer MLP, maps the higher dimensional latent distribution back to the input distribution space. The architectures for the two MLPs are shown in Table~\ref{tab:mlp}. 

\begin{table}
\begin{center}
\caption{Network Architectures of the encoder and the decoder for the synthetic experiments to validate the distribution alignment mechanism of the proposed regularization. FC-x represents fully-connected layer with x hidden neurons. ReLU denotes the ReLU activation. BatchNorm represents the batch normalization.}
\label{tab:mlp}
\small
\begin{tabular}{c | c}
\hline
\textbf{Model}&\textbf{Architecture}\\
\hline
Encoder \textbf{G} & FC-56, ReLU, FC-128, ReLU,\\
 & FC-256, ReLU, BatchNorm\\
\hline
Decoder \textbf{D} & FC-128, ReLU, BatchNorm,\\
 & FC-56, ReLU, FC-2, ReLU\\
\hline
\end{tabular}
\end{center}
\end{table}

\begin{figure}[h]
    \center
    \begin{subfigure}[b]{1.0\linewidth}
        \includegraphics[width=1.0\linewidth]{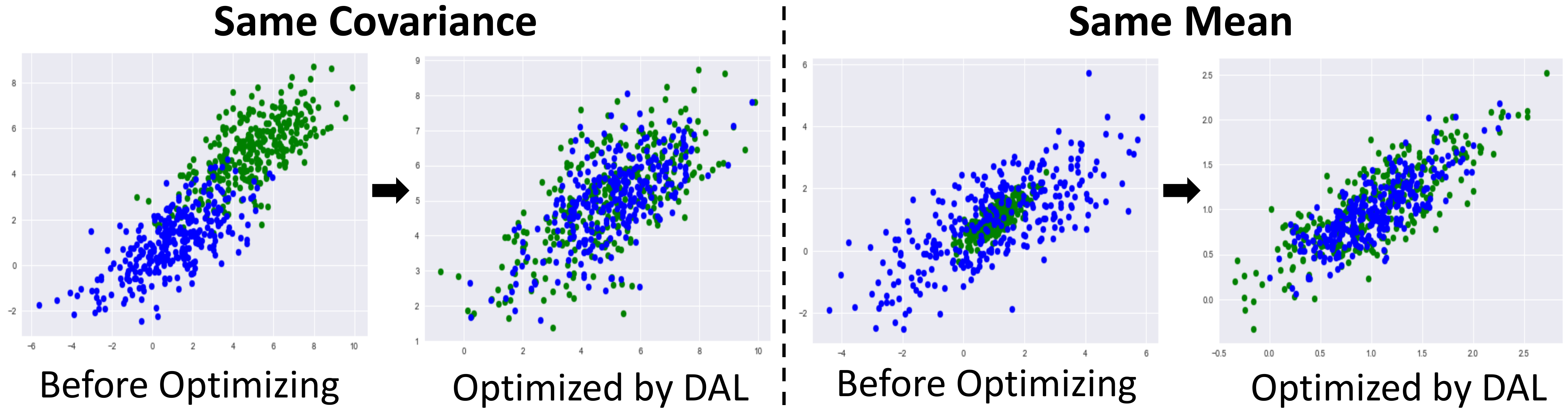}
        \caption{2D Gaussian.}
    \end{subfigure}
    \begin{subfigure}[b]{0.49\linewidth}
        \includegraphics[width=1.0\linewidth]{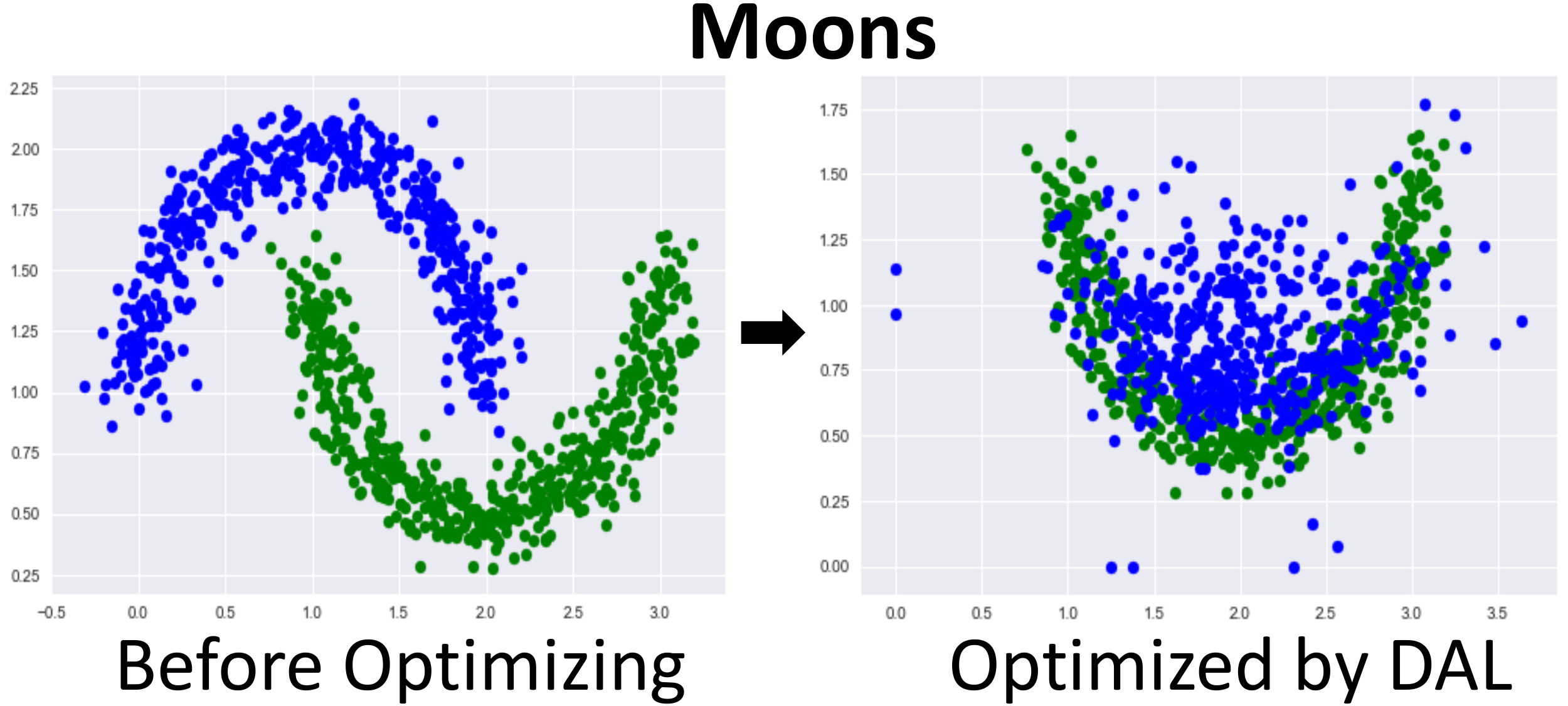}
        \caption{Moons.}
    \end{subfigure}
    \begin{subfigure}[b]{0.49\linewidth}
        \includegraphics[width=1.0\linewidth]{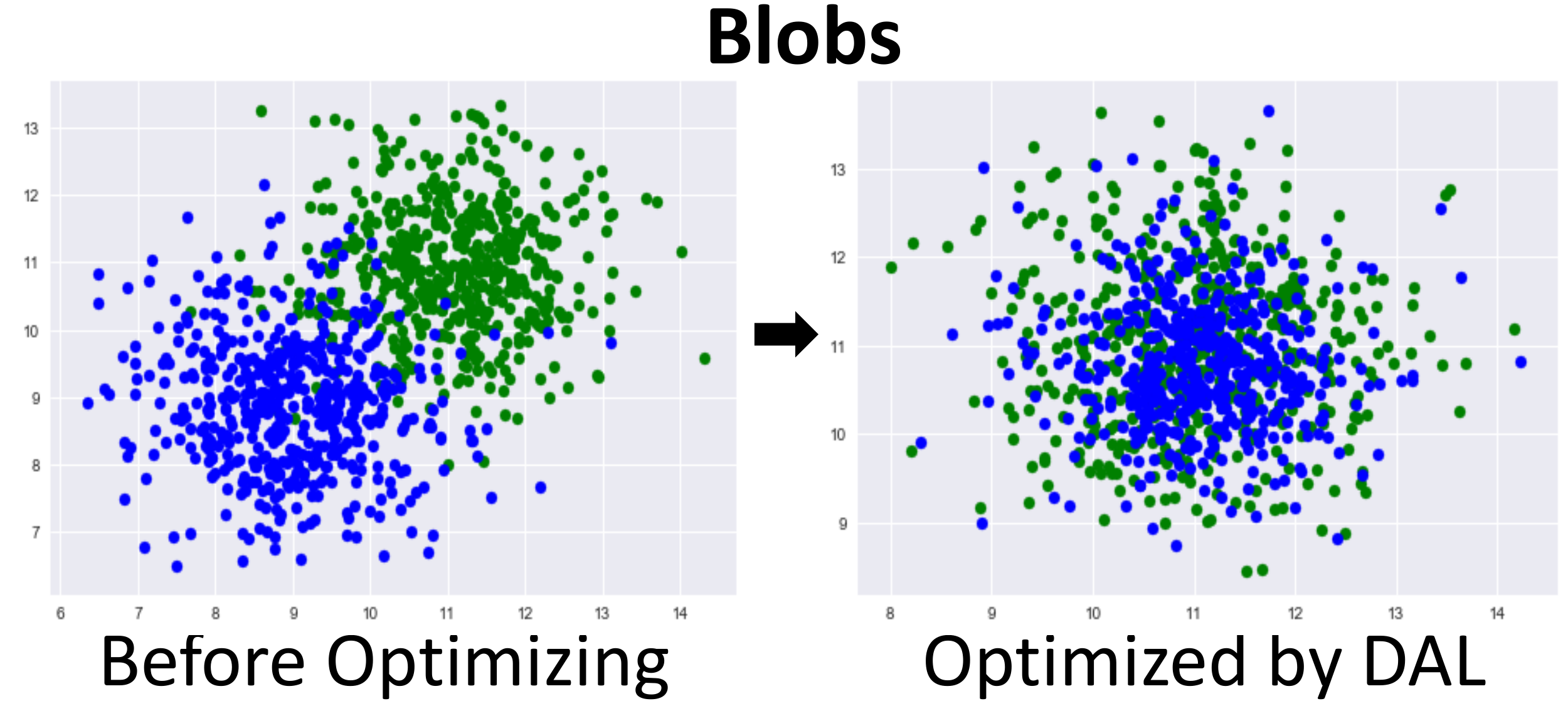}
        \caption{Blobs.}
    \end{subfigure}
    \caption{\textbf{(Best viewed in color.)} \textbf{Green} and \textbf{blue} points indicate the samples from the source distribution and the target distribution, respectively. The predicted target distribution well aligns with the source distribution after the proposed \emph{distribution alignment loss} converges, which validates the distribution alignment mechanism of \emph{distribution alignment loss}.}\label{fig:syn}
\end{figure}

The samples from the target input distribution are fed into the encoder $G$ and the decoder $D$ to generate their predictions $D(G(x_t))$. The outputs of $D$, which are the predicted target samples, and the samples from the source input distribution are utilized for the proposed DAL. We tested the same covariance case and the same mean case for the 2D Gaussian distributions. For the same covariance case, the \textbf{green} points (source) were sampled from a 2D Gaussian with mean $\begin{pmatrix}
5&5\\
\end{pmatrix}$ and covariance $\begin{pmatrix}
4&2\\
2&2\\
\end{pmatrix}$; and the \textbf{blue} points (target) indicate the samples from a 2D Gaussian with the same covariance but different mean $\begin{pmatrix}
1&1\\
\end{pmatrix}$. For the same mean case, the two 2D Gaussian distributions have the same mean $\begin{pmatrix}
1&1\\
\end{pmatrix}$ but different covariance, i.e., $\begin{pmatrix}
0.3&0.2\\
0.2&0.2\\
\end{pmatrix}$ for the source input distribution and 
$\begin{pmatrix}
4&2\\
2&2\\
\end{pmatrix}$
 for the target input distribution. We used \emph{scikit-learn} \cite{scikit} to generate \emph{moons} and \emph{blobs} datasets. For \emph{moons} dataset, we made two interleaving half circles for the two domains and add a Gaussian noise with standard deviation 0.1 to the data. For \emph{blobs} dataset, we generated two isotropic Gaussian blobs with centers at $\begin{pmatrix}
11&11\\
\end{pmatrix}$ and 
$\begin{pmatrix}
9&9\\
\end{pmatrix}$ for the source input distribution and the target input distribution, respectively. As shown in Figure~\ref{fig:syn}, the predicted target samples (\textbf{blue} points) successfully align with the source samples (\textbf{green} points) after optimizing by DAL alone in all synthetic experiments. Therefore, we can claim that the proposed DAL serves as the distribution alignment mechanism in our model.

\subsection{Digit Classification}
\subsubsection{Setup}
In this section, we evaluated the adaptation of our two frameworks DFA-ENT and DFA-MCD on five digit and traffic-sign recognition datasets. For each adaptation scenario, we employed the same network architectures utilized in \cite{pixelDA,dann,saito}, and implemented the decoder $D$ accordingly. To evaluate DFA-ENT, we used the SGD optimizer with a mini-batch size of 256 in all digit and traffic-sign recognition experiments. We set the learning rate to $0.1$ in the adaptation from SVHN to MNIST and $0.02$ in other adaptation scenarios for evaluating DFA-ENT. Our hyper-parameters $\alpha$ and $\beta$ were set to $0.01$ and $10$, respectively, in all adaptation scenarios for both frameworks.

\textbf{SVHN (SV)} $\rightarrow$ \textbf{MNIST (MN)}: Street-View House Number (SVHN) \cite{svhn} and MNIST \cite{mnist} datasets were used as the source domain and the target domain, respectively. The two datasets consist of images of digit from 0 to 9. However, SVHN \cite{svhn} has significant variations in the colored background, contrast, rotation, scale, etc.

\textbf{MNIST (MN)} $\leftrightarrow$ \textbf{USPS (US)}: We evaluated two adaptation scenarios on USPS \cite{usps} and MNIST \cite{mnist} datasets. We used the same setup provided by \cite{saito} for the two adaptation scenarios.

\textbf{SYN SIGNS (SY)} $\rightarrow$ \textbf{GTSRB (GT)}: We also evaluated the proposed frameworks on a more complex scenario, from synthetic traffic signs dataset (SYN SIGNS) \cite{synsig} to the real-world German Traffic Signs Recognition Benchmark (GTSRB) \cite{gts}. This domain adaptation scenario has 43 different traffic signs (classes). We split the datasets based on \cite{saito}.

\subsubsection{Results}
\begin{table}[h]
\setlength{\tabcolsep}{1pt}
\setlength\aboverulesep{0pt}\setlength\belowrulesep{0pt}
\caption{Accuracy(\%) of the proposed frameworks on the benchmark datasets for digit and traffic-sign recognition.}
\label{tab:mcd}
\small
\begin{center}
\begin{tabular}{c c c c c c}
\hline
\textbf{Method}&\textbf{SV}$\veryshortarrow$\textbf{MN}&\textbf{SY}$\veryshortarrow$\textbf{GT}&\textbf{MN}$\veryshortarrow$\textbf{US}&\textbf{$\textrm{MN}^\ast$}$\veryshortarrow$\textbf{$\textrm{US}^\ast$} & \textbf{US}$\veryshortarrow$\textbf{MN} \\

\hline
\hline
Source Only & 67.1 & 85.1 & 76.7 & 79.4& 63.4 \\
\hline
DANN\cite{dann} & 71.1 & 88.7 & 77.1 & 85.1 & 73.0\\
DSN\cite{dsn} & 82.7 & 93.1 & 91.3 &-& - \\
ADDA\cite{adda} & 76.0 & - & 89.4 &-& 90.1 \\
MSTN\cite{mstn} & 91.7 & - & - & 92.9 & - \\
GTA\cite{gta} & 92.4 & - & 92.8 & 95.3& 90.8 \\
DEV\cite{dev} & 93.2 & - & - & 92.5 & \textbf{96.9} \\
GPDA\footnotemark \cite{kim2019unsupervised} & 98.2 & 96.2 & 96.4 & 98.1 & 96.4\\
\hline
MCD\cite{saito} & 96.2 & 94.4 & 94.2 & 96.5 & 94.1\\
(n = 4)& $\pm$ 0.4 & $\pm$ 0.3 & $\pm$ 0.7 & $\pm$ 0.3& $\pm$ 0.3 \\
\hline
\hline
\textbf{DFA-ENT} & 98.2 & 96.8 & 96.5 & 97.9 & 96.2 \\
\textbf{ (Ours)} &  $\pm$ 0.3 &  $\pm$ 0.2 &  $\pm$ 0.4 & $\pm$ 0.2 & $\pm$ 0.1 \\
\hline
\textbf{DFA-MCD} & \textbf{98.9} & \textbf{97.5} & \textbf{97.3} & \textbf{98.6} & 96.6 \\
\textbf{ (Ours)} &  $\pm$ 0.2 &  $\pm$ 0.2 &  $\pm$ 0.1 & $\pm$ 0.1 & $\pm$ 0.2 \\
\hline
\end{tabular}
\end{center}
\end{table}

\footnotetext{This framework is developed based on MCD.}

\begin{figure}[t]
\begin{center}
\includegraphics[width=1.0\linewidth]{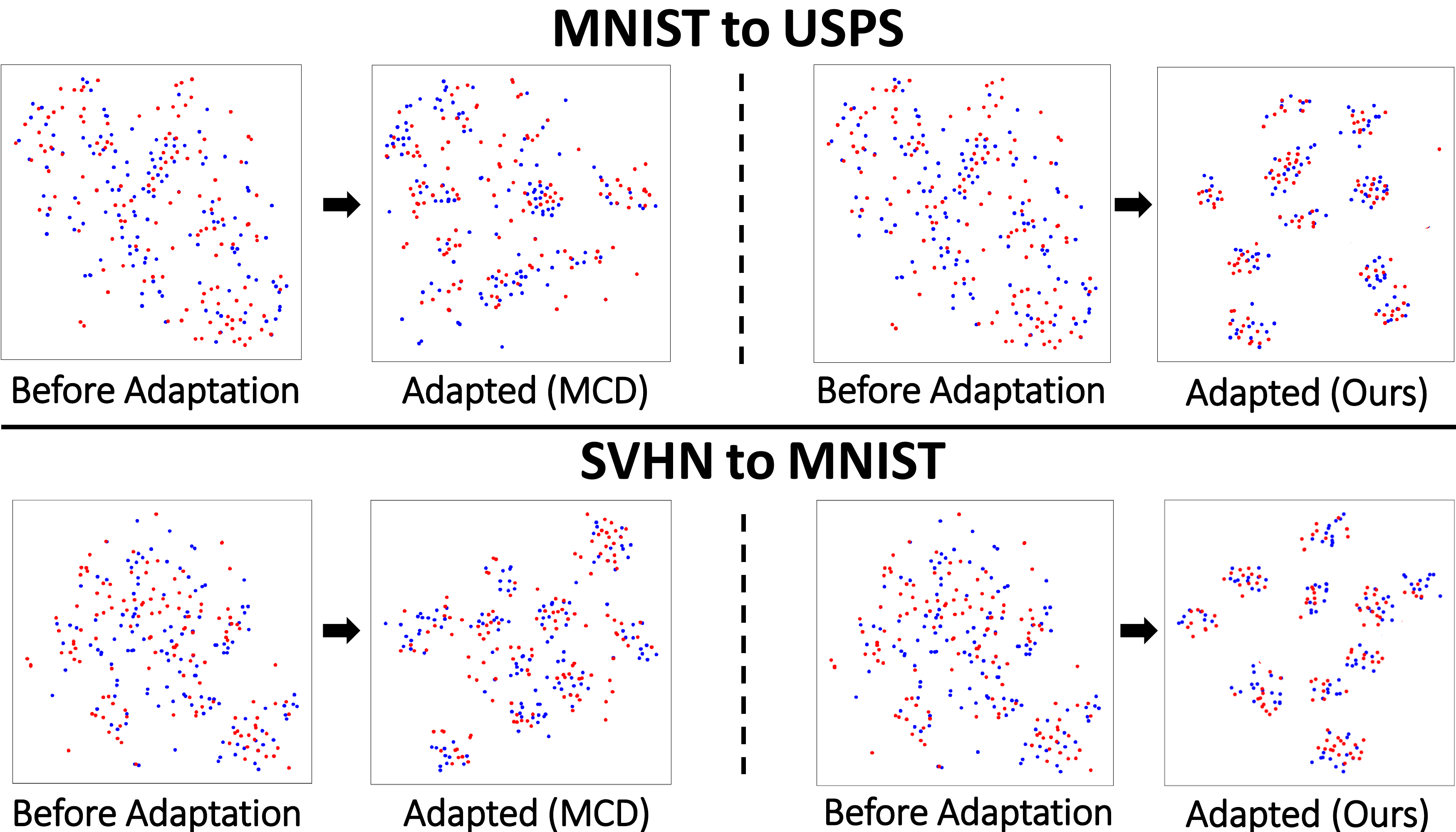}
\end{center}
   \caption{\textbf{(Best viewed in color.)} t-SNE \cite{tsne} visualizations of the learned feature representations for two different adaptation scenarios. \textbf{Blue} and \textbf{red} points indicate the latent features from the source domain and the target domain, respectively.}
\label{fig:tsne}
\end{figure}

Table~\ref{tab:mcd} lists the results for the target domain classification. $\{dataset\}^{\ast}$ denotes that all of the training samples are used for training the frameworks. We used the same networks for the source only evaluation. The average and the standard deviation of the accuracy on each DA scenario are reported by repeating each experiment 5 times. The results indicate that our model significantly improves the adaptation performance of MCD on all digit and traffic-sign datasets. The standard deviations of DFA-MCD are much lower than those of MCD, which indicates that our model can result in more robust performance. The visualizations of the learned feature representations are shown in Figure~\ref{fig:tsne}. The comparison is conducted between DFA-MCD and MCD. The better feature clustering indicates that our model significantly improves the adaptation performance of MCD through better feature alignment.

\subsection{Object Classification}
\subsubsection{Setup}
We extensively evaluated the adaptation performance of DFA-ENT and DFA-SAFN on five benchmark datasets for object recognition, namely, \emph{VisDA2017}, \emph{Office-31}, \emph{ImageCLEF-DA} and \emph{Office-Home}. For each adaptation scenario, we employed ResNet-50 \cite{resnet} that was fine-tuned from the ImageNet \cite{imagenet} pre-trained model. We implemented our decoder $D$ as an inverted version of the feature extractor of ResNet-50. To evaluate DFA-ENT, we used the SGD optimizer with a learning rate of $1 \times 10^{-3}$, and set the batch size to 32 on all benchmark datasets. Our hyper-parameters $\alpha$ and $\beta$ were set to 0.1 and 10, respectively, for both frameworks. 

\textbf{VisDA2017} \cite{peng2017visda} is a large-scale benchmark dataset used for the 2017 visual domain adaptation challenge. The goal of the dataset is trying to bridge the domain gap between the synthetic objects and the real obbjects. It has over 280K images across 12 object categories. The source domain consists of 152,397 synthetic images that are generated by rendering the 3D models of a certain object categories. The target domain contains 55,388 images of the real objects, which are collected from Microsoft COCO dataset \cite{lin2014microsoft}. This could be the most challenging benchmark dataset for UDA.

\begin{figure}[h]
    \center
    \begin{subfigure}[b]{0.45\linewidth}
        \includegraphics[width=1.0\linewidth]{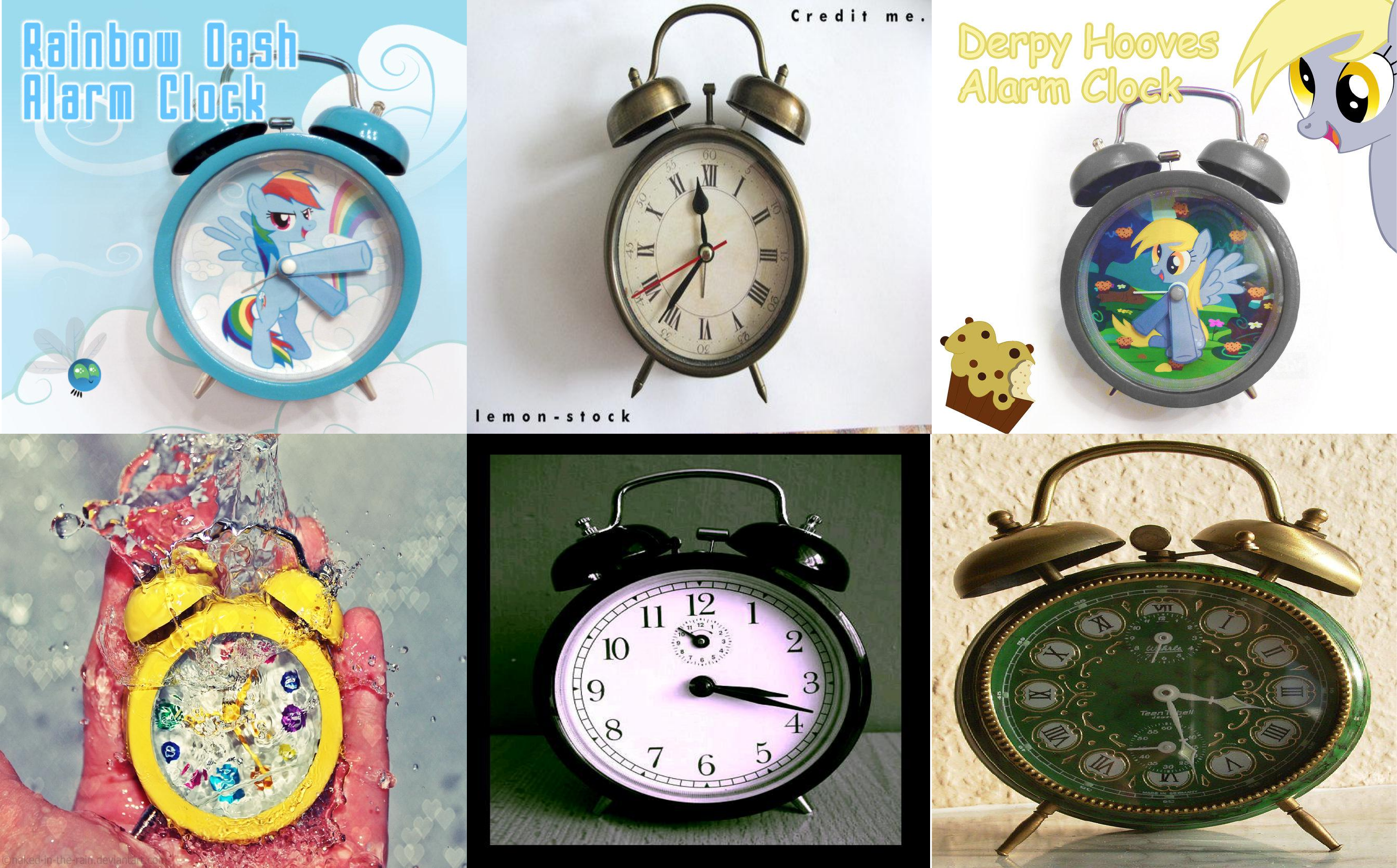}
        \caption{Artistic.}
    \end{subfigure}
    \begin{subfigure}[b]{0.45\linewidth}
        \includegraphics[width=1.0\linewidth]{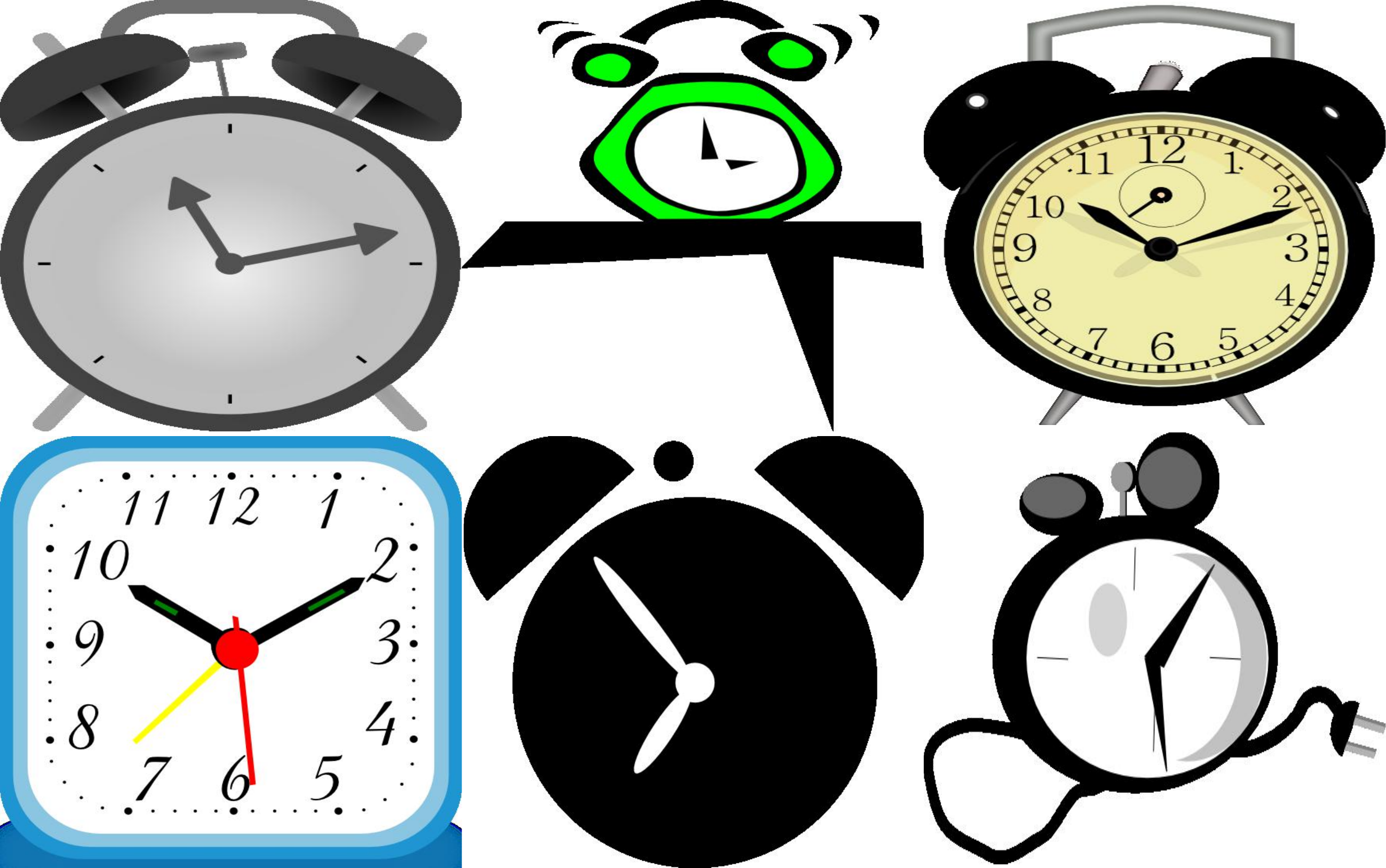}
        \caption{Clipart.}
    \end{subfigure}
    \begin{subfigure}[b]{0.45\linewidth}
        \includegraphics[width=1.0\linewidth]{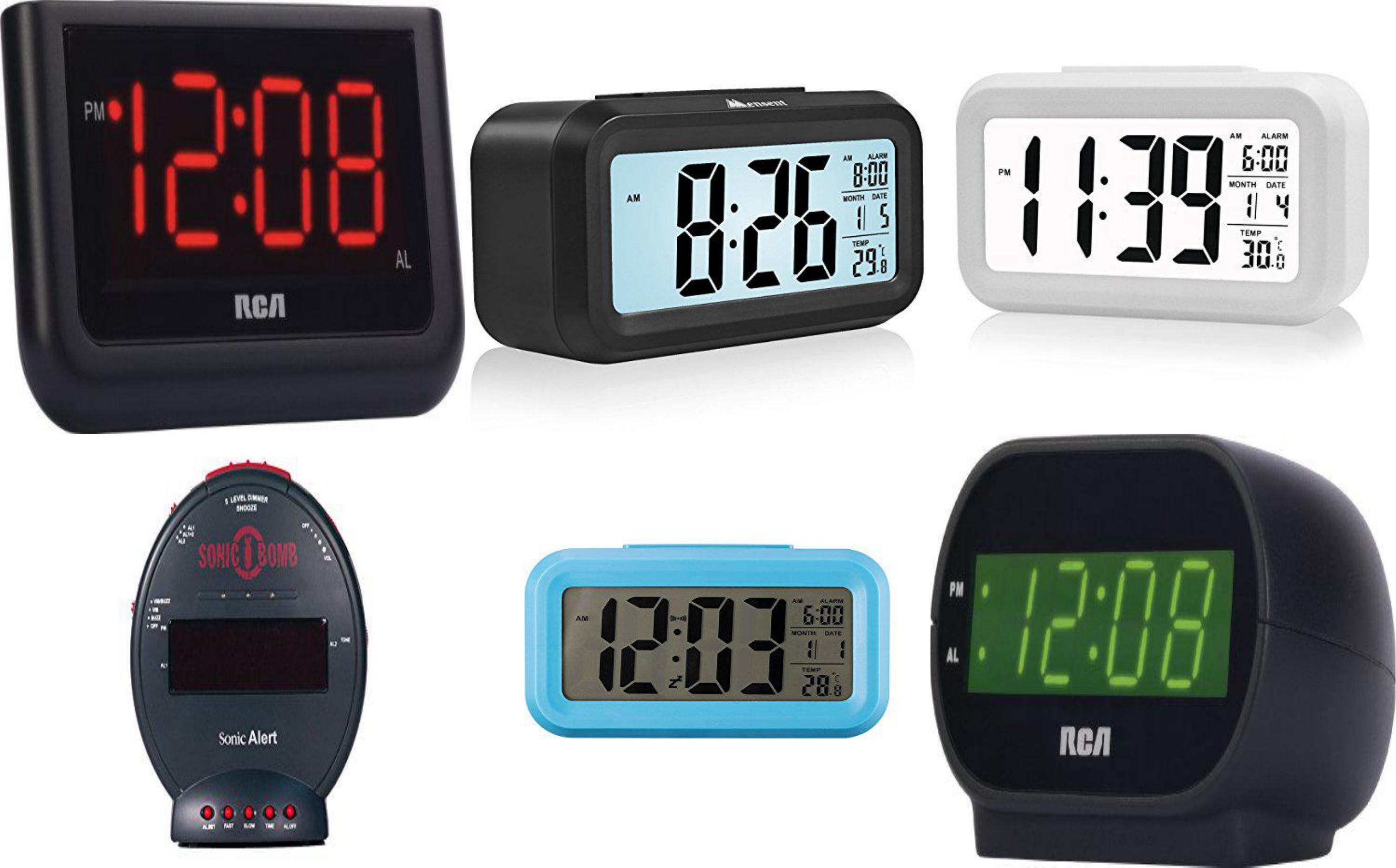}
        \caption{Product.}
    \end{subfigure}
    \begin{subfigure}[b]{0.45\linewidth}
        \includegraphics[width=1.0\linewidth]{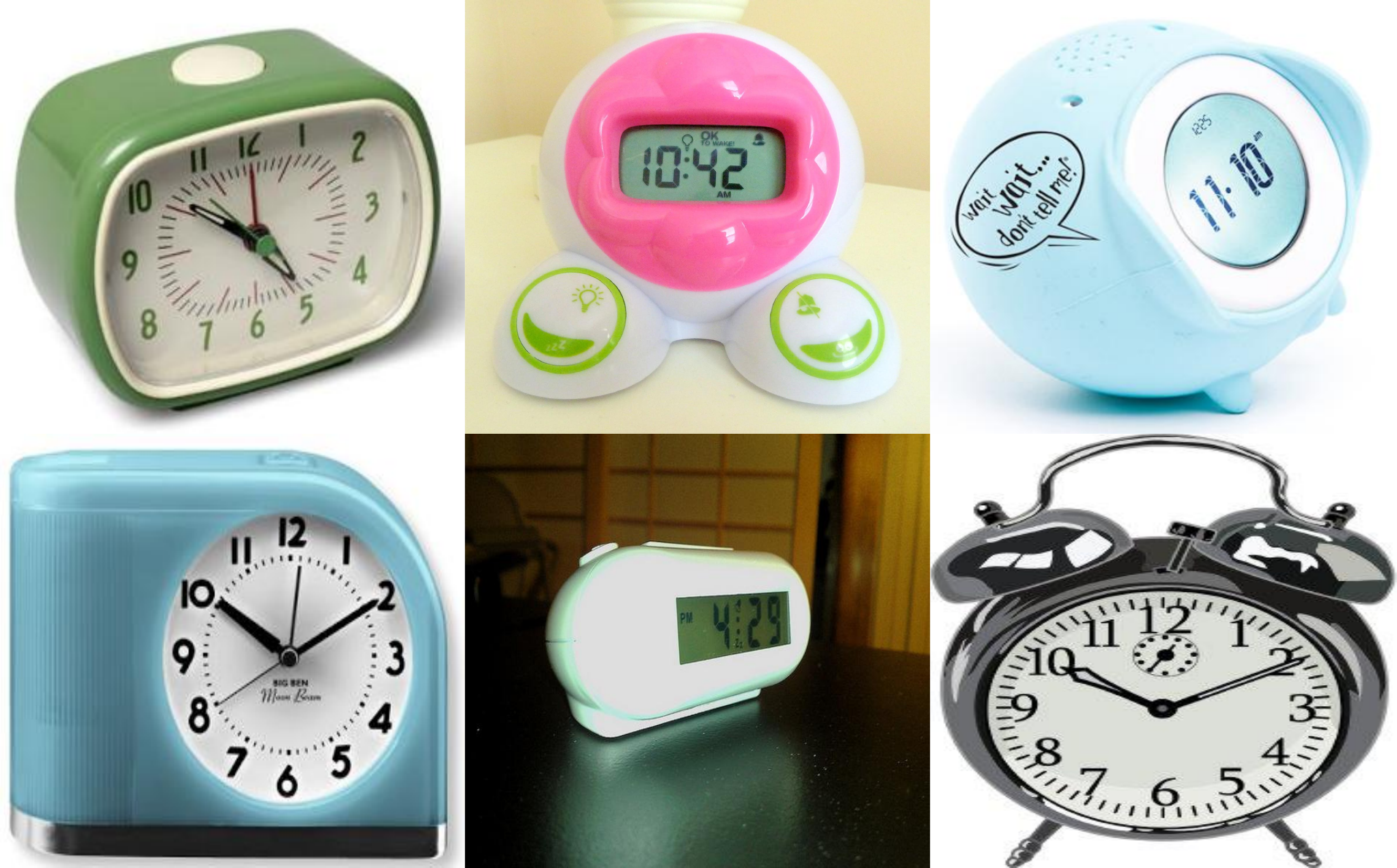}
        \caption{Real-World.}
    \end{subfigure}
    \caption{Example images for alarm clock from the four different domains of Office-Home. }\label{fig:office-home}
\end{figure}

\textbf{Office-Home} \cite{officehome} has images of everyday objects from four different domains: Artistic (\textbf{Ar}), Clipart (\textbf{Cl}), Product (\textbf{Pr}) and Real-World (\textbf{Rw}). The dataset has around 15,500 images. Each domain contains 65 object classes. Notably, \textbf{Ar} consists of the images from the different forms of artistic depictions of objects, while a regular camera takes the images of \textbf{Rw}. Some image samples from this dataset are shown in Figure~\ref{fig:office-home}.

\begin{table*}
\setlength{\tabcolsep}{4.4pt}
\caption{Accuracy(\%) of the proposed frameworks on \emph{VisDA2017} (ResNet-50).}
\label{tab:visda2017}
\small
\begin{center}
\begin{tabular}{c c c c c c c c c c c c c | c}
\hline
\textbf{Method} & \textbf{plane} & \textbf{bcycl} & \textbf{bus} & \textbf{car} & \textbf{horse} & \textbf{knife} & \textbf{mcycl} & \textbf{person} & \textbf{plant} & \textbf{sktbrd} & \textbf{train} & \textbf{truck} & \textbf{Per-class} \\
\hline
\hline
ResNet-50 \cite{resnet} & 60.2 & 10.3 & 54.7 & 54.5 & 42.9 & 2.1 & 78.9 & 4.5 & 45.5 & 29.5 & 89.0 & 12.4 & 40.4\\
SAFN \cite{norm} & 90.5 & 55.9 & 80.3 & 64.6 & 88.8 & 31.8 & 92.7 & 70.4 & \textbf{93.2} & 49.6 & 87.7 & 23.2 & 69.1 \\
MCD \cite{saito} & 90.3 & 62.6 & 84.8 & 71.7 & 85.9 & 72.9 & \textbf{93.7} & 71.9 & 86.8 & 79.1 & 81.6 & 14.3 & 74.6\\
\hline
\hline
\textbf{DFA-ENT (Ours)} & 88.3 & 55.1 & 81.0 & \textbf{72.9} & \textbf{91.4} & 94.4 & 91.1 & 75.1 & 80.6 & 45.7 & 88.2 & 15.8 & 73.3\\
\hline
\textbf{DFA-SAFN (Ours)} & \textbf{93.1} & 58.4 & \textbf{85.8} & 69.9 & 89.8 & \textbf{96.1} & 90.3 & 77.5 & 87.4 & 48.9 & 85.1 & 21.1 & 75.3\\
\textbf{DFA-MCD (Ours)} & 91.2 & \textbf{77.4} & 80.5 & 63.3 & 87.1 & 85.4 & 86.4 & \textbf{79.5} & 90.3 & \textbf{79.7} & \textbf{89.2} & \textbf{31.6} & \textbf{78.5}\\
\hline
\end{tabular}
\end{center}
\end{table*}

\textbf{ImageCLEF-DA}\footnote{https://www.imageclef.org/2014/adaptation} is a dataset used for the 2014 ImageCLEF domain adaptation challenge. This dataset selects 12 common object classes from three public datasets: \emph{Caltech-256} (\textbf{C}), \emph{ImageNet ILSVRC2012} (\textbf{I}) and \emph{Pascal VOC 2012} (\textbf{P}). The dataset organizers selected 50 images per class and 600 images in total for each domain.

\textbf{Office-31} \cite{office31} is a standard benchmark dataset for evaluating visual DA algorithms. It has three different domains: \emph{Amazon} (\textbf{A}), \emph{Webcam} (\textbf{W}), and \emph{DSLR} (\textbf{D}). \emph{Amazon} consists of images from amazon.com. \emph{Webcam} and \emph{DSLR} contain images for the office environment captured by a web camera and a digital SLR camera, respectively. It consists of 4,652 images of 31 object categories.

\subsubsection{Results}
\begin{table}[h]
\setlength{\tabcolsep}{1.5pt}
\caption{Accuracy(\%) of the proposed frameworks on \emph{ImageCLEF-DA} (ResNet-50).}
\label{tab:imageclef}
\small
\begin{center}
\begin{tabular}{c c c c c c c | c}
\hline
\textbf{Method}& \textbf{I$\rightarrow$P} & \textbf{P$\rightarrow$I} & \textbf{I$\rightarrow$C} & \textbf{C$\rightarrow$I} & \textbf{C$\rightarrow$P} & \textbf{P$\rightarrow$C} & \textbf{Avg}\\
\hline
\hline
ResNet-50 \cite{resnet} & 74.8 & 83.9 & 91.5 & 78.0 & 65.5 & 91.2 & 80.7\\
\hline
DANN \cite{dann} & 75.0 & 86.0 & 96.2 & 87.0 & 74.3 & 91.5 & 85.0 \\
CDAN$^\ast$\cite{CADA2018Long} & 76.7 & 90.6 & 97.0 & 90.5 & 74.5 & 93.5 & 87.1\\
CADA\cite{kurmi2019attending} & 78.0 & 90.5 & 96.7 & 92.0 & 77.2 & 95.5 & 88.3\\
CDAN+TN \cite{wang2019transferable}& 78.3 &	90.8 & 96.7& 92.3 & 78.0 & 94.8 & 88.5\\
HAFN \cite{norm}& 76.9 & 89.0 & 94.4 & 89.6 & 74.9 & 92.9 & 86.3\\
\hline
SAFN \cite{norm} & 79.3 & 93.3 & 96.3 & 91.7 & 77.6 & 95.3 & 88.9\\
& $\pm$ 0.1 & $\pm$ 0.4 &$\pm$ 0.4& $\pm$ 0.0 & $\pm$ 0.1&$\pm$ 0.1 & \\
\hline
\hline
\textbf{DFA-ENT} & 79.5 & 93.0 & 96.4 & 92.5 & 77.2 & 95.8 & 89.1\\
\textbf{(Ours)} & $\pm$ 0.0 & $\pm$ 0.3 & $\pm$ 0.2 & $\pm$ 0.2 & $\pm$ 0.1 & $\pm$ 0.3 & \\
\hline
\textbf{DFA-SAFN} & \textbf{80.0} & \textbf{94.2} & \textbf{97.5}& \textbf{93.8} & \textbf{78.7} & \textbf{96.7} & \textbf{90.2}\\
\textbf{(Ours)} & $\pm$ 0.1& $\pm$ 0.3 & $\pm$ 0.2 & $\pm$ 0.0 & $\pm$ 0.1 & $\pm$ 0.0 & \\
\hline
\end{tabular}
\end{center}
\end{table}

\begin{table}[h]
\setlength{\tabcolsep}{1.5pt}
\caption{Accuracy(\%) of the proposed frameworks on \emph{Office-31} (ResNet-50).}
\label{tab:office31}
\small
\begin{center}
\begin{tabular}{c c c c c c c | c}
\hline
\textbf{Method}& \textbf{A$\rightarrow$W} & \textbf{D$\rightarrow$W} & \textbf{W$\rightarrow$D} & \textbf{A$\rightarrow$D} & \textbf{D$\rightarrow$A} & \textbf{W$\rightarrow$A} & \textbf{Avg}\\
\hline
\hline
ResNet-50 \cite{resnet} & 68.4 & 96.7 & 99.3 & 68.9& 62.5 &60.7 & 76.1\\
\hline
DANN \cite{dann} & 82.0 & 96.9 & 99.1 & 79.7& 68.2 & 67.4& 82.2 \\
GTA \cite{gta} & 89.5 & 97.9 & 99.8 & 87.7& 72.8 & 71.4 & 86.5 \\
CDAN$^\ast$\cite{CADA2018Long} & 93.1 & 98.2 & 100.0 & 89.8 & 70.1 & 68.0 & 86.6\\
DSBN\cite{chang2019domain} & 93.3 & 99.1 & 100.0 & 90.8 & 72.7 & \textbf{73.9} & 88.3\\
TAT\cite{tat} & 92.5 & 99.3 & 100.0 & 93.2 & 73.1 & 72.1 & 88.4\\
HAFN \cite{norm}& 83.4 & 98.3 & 99.7 & 84.4 & 69.4 & 68.5 & 83.9\\
\hline
SAFN \cite{norm} & 90.1 & 98.6 & 99.8 & 90.7 & 73.0 & 70.2 & 87.1\\
& $\pm$ 0.8 & $\pm$ 0.2 &$\pm$ 0.0& $\pm$ 0.5 & $\pm$ 0.2&$\pm$ 0.3 & \\
\hline
\hline
\textbf{DFA-ENT} & 90.5 & 99.0 & 100.0 & 94.3 & 72.1 & 67.8 & 87.3\\
\textbf{(Ours)} & $\pm$ 0.7 & $\pm$ 0.1 & $\pm$ 0.0 & $\pm$ 0.4 & $\pm$ 0.2 & $\pm$ 0.4 & \\
\hline
\textbf{DFA-SAFN} & \textbf{93.5} & \textbf{99.4}& \textbf{100.0}& \textbf{94.8} & \textbf{73.8} & 71.0 & \textbf{88.8}\\
\textbf{(Ours)} & $\pm$ 0.5 & $\pm$ 0.1 & $\pm$ 0.0 & $\pm$ 0.3 & $\pm$ 0.1 & $\pm$ 0.2& \\
\hline
\end{tabular}
\end{center}
\end{table}

\begin{table*}
\setlength{\tabcolsep}{4.4pt}
\caption{Accuracy(\%) of the proposed frameworks on \emph{Office-Home} (ResNet-50).}
\label{tab:officehome}
\small
\begin{center}
\begin{tabular}{c c c c c c c c c c c c c | c}
\hline
\textbf{Method} & \textbf{Ar}$\veryshortarrow$\textbf{Cl} & \textbf{Ar}$\veryshortarrow$\textbf{Pr} & \textbf{Ar}$\veryshortarrow$\textbf{Rw} & \textbf{Cl}$\veryshortarrow$\textbf{Ar} & \textbf{Cl}$\veryshortarrow$\textbf{Pr} & \textbf{Cl}$\veryshortarrow$\textbf{Rw} & \textbf{Pr}$\veryshortarrow$\textbf{Ar} & \textbf{Pr}$\veryshortarrow$\textbf{Cl} & \textbf{Pr}$\veryshortarrow$\textbf{Rw} & \textbf{Rw}$\veryshortarrow$\textbf{Ar} & \textbf{Rw}$\veryshortarrow$\textbf{Cl} & \textbf{Rw}$\veryshortarrow$\textbf{Pr} & \textbf{Avg} \\
\hline
\hline
ResNet-50 \cite{resnet} & 34.9 & 50.0 & 58.0 & 37.4 & 41.9 & 46.2 & 38.5 & 31.2 & 60.4 & 53.9 & 41.2 & 59.9 & 46.1\\
\hline
DANN\cite{dann} & 45.6 & 59.3 & 70.1 & 47.0 & 58.5 & 60.9 & 46.1 & 43.7 & 68.5 & 63.2 & 51.8 & 76.8 & 57.6 \\
CDAN$^\ast$\cite{CADA2018Long} & 49.0 & 69.3 & 74.5 & 54.4 & 66.0 & 68.4 & 55.6 & 48.3 & 75.9 & 68.4 & 55.4 & 80.5 & 63.8\\
DWT-MEC\cite{roy2019unsupervised} & 50.3 & 72.1 & 77.0 & 59.6 & 69.3 & 70.2 & 58.3 & 48.1 & 77.3 & 69.3 & 53.6 & 82.0 & 65.6\\
TAT\cite{tat} & 51.6 & 69.5& 75.4& 59.4& 69.5& 68.6& 59.5& 50.5& 76.8& 70.9& 56.6& 81.6& 65.8\\
CDAN+TN \cite{wang2019transferable} &50.2& 71.4& 77.4& 59.3& 72.7& 73.1& 61.0&	\textbf{53.1}& 79.5& 71.9& \textbf{59.0}& 82.9 & 67.6\\
HAFN \cite{norm} & 50.2 & 70.1 & 76.6 & 61.1 & 68.0 & 70.7 & 59.5 & 48.4 & 77.3 & 69.4 & 53.0 & 80.2 & 65.4\\
\hline
SAFN \cite{norm} & 52.0 & 71.7 & 76.3 & 64.2 & 69.9 & 71.9 & 63.7 & 51.4 & 77.1 & 70.9& 57.1 & 81.5 & 67.3\\
& $\pm$ 0.1 & $\pm$ 0.6 & $\pm$ 0.3 & $\pm$ 0.3 & $\pm$ 0.6 & $\pm$ 0.6 & $\pm$ 0.4 & $\pm$ 0.2 & $\pm$ 0.0 & $\pm$ 0.4 & $\pm$ 0.1 & $\pm$ 0.0 & \\
\hline
\hline
\textbf{DFA-ENT} & 50.6 & \textbf{74.8} & \textbf{79.3} & 65.2 & \textbf{73.8} & \textbf{74.5} & 63.5 & 51.4 & \textbf{81.4} & 73.9 & 58.2 & \textbf{83.3} & \textbf{69.2}\\
\textbf{(Ours)}& $\pm$ 0.1 & $\pm$ 0.3 & $\pm$ 0.2 & $\pm$ 0.2 & $\pm$ 0.3 & $\pm$ 0.4 & $\pm$ 0.4 & $\pm$ 0.3 & $\pm$ 0.0 & $\pm$ 0.4 & $\pm$ 0.0 & $\pm$ 0.0 & \\
\hline
\textbf{DFA-SAFN} & \textbf{52.8} & 73.9 & 77.4 & \textbf{66.5} & 72.9 & 73.6 & \textbf{64.9} & \textbf{53.1} & 78.7 & \textbf{74.5} & 58.1 & 82.4 & 69.1\\
\textbf{(Ours)}& $\pm$ 0.1 & $\pm$ 0.4 & $\pm$ 0.2 & $\pm$ 0.1 & $\pm$ 0.3 & $\pm$ 0.3 & $\pm$ 0.2 & $\pm$ 0.1 & $\pm$ 0.0 & $\pm$ 0.3 & $\pm$ 0.0 & $\pm$ 0.0 & \\
\hline
\end{tabular}
\end{center}
\end{table*}

The results of DFA-ENT and DFA-SAFN on \emph{VisDA2017}, \emph{ImageCLEF-DA}, \emph{Office-31} and \emph{Office-Home} are listed in Table~\ref{tab:visda2017}, \ref{tab:imageclef}, \ref{tab:office31} and \ref{tab:officehome}, respectively. \emph{\{Method}\}$^\ast$ indicates that ten-crop images are used in the evaluation phase with its best-performing models. We repeated each experiment 3 times and reported the average and the standard deviation of the accuracy for evaluating the datasets \emph{Office-Home}, \emph{ImageCLEF-DA} and \emph{Office-31}. We reported the accuracy of the evaluation on \emph{VisDA2017} after 20 epochs with no repeated experiments. The results illustrate that the proposed frameworks significantly outperform the benchmark algorithms on object classification. The robustness of SAFN is also improved by DFA with lower variance among each repeated experiments. 

Results on \emph{VisDA2017} show that the proposed DFA can significantly help the existing methods to better bridge the synthetic-to-real domain gap, which improves the performance of the baseline methods by at least 3.9\% (6.2\% for SAFN and 3.9\% for MCD). Notably, the proposed DFA-MCD achieves state-of-the-art performance on this large-scale dataset. Besides, our simplified framework DFA-ENT achieves the competitive performance in all four beachmark datasets for the object recognition task, which suggests the effectiveness of the latent alignment in transfer learning. Moreover, the outstanding improvement on the adaptation scenarios (\emph{Office-31}, \emph{Office-Home}) with significant nuisance image variations suggests that our model can improve other frameworks' knowledge transferability remarkably in the adaptation scenario with significant variations.

One interesting observation can be revealed from these results that the transfer gains of the existing approaches, which mitigate the domain gap by classifier-induced discrepancies, can be further improved by improving the alignment in the feature spaces. One limitation of our research is that we only consider the way to better construct the feature spaces for the DA problem and directly incorporate the proposed method into the classifier-induced discrepancy based methods. Therefore, we believe that the transfer gains can be more significantly improved by explicitly considering the relationship between the features induced from the feature extractor and the feature induced from the classifier. But how to trade off the alignment of the latent distributions against the alignment of the output class distributions is still a big challenge for the DA community.


\section{Ablation Study}\label{sec:ablation}

\subsection{The Shape of The Latent Distribution} \label{sec:ab_shape}
\begin{figure}[h]
    \center
    \begin{subfigure}[b]{0.49\linewidth}
        \includegraphics[width=1.0\linewidth]{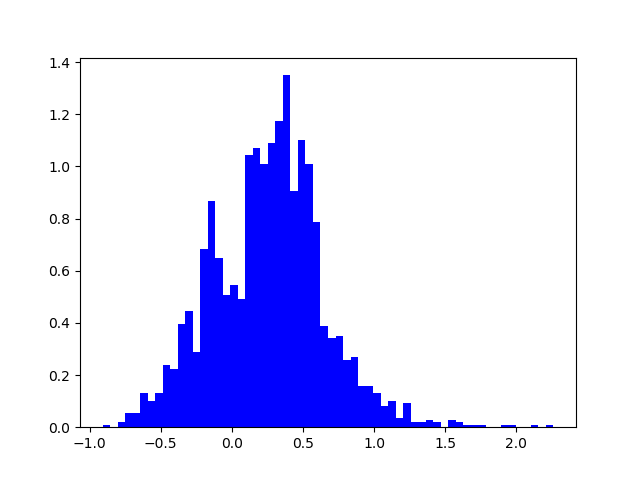}
        \caption{Source Latent (Office-31).}
    \end{subfigure}
    \begin{subfigure}[b]{0.49\linewidth}
        \includegraphics[width=1.0\linewidth]{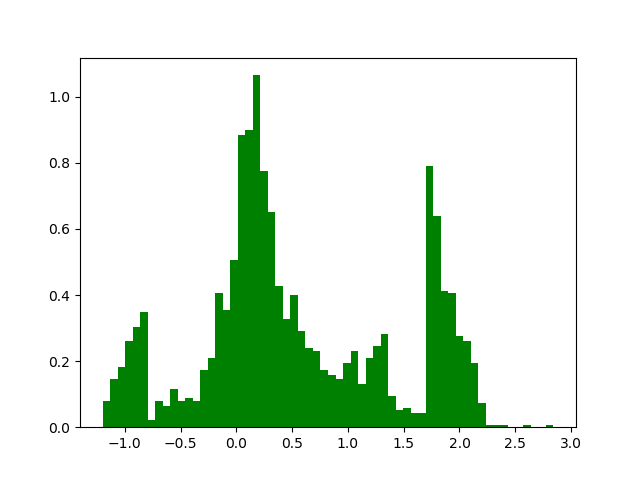}
        \caption{Target Latent (Office-31).}
    \end{subfigure}
    \begin{subfigure}[b]{0.49\linewidth}
        \includegraphics[width=1.0\linewidth]{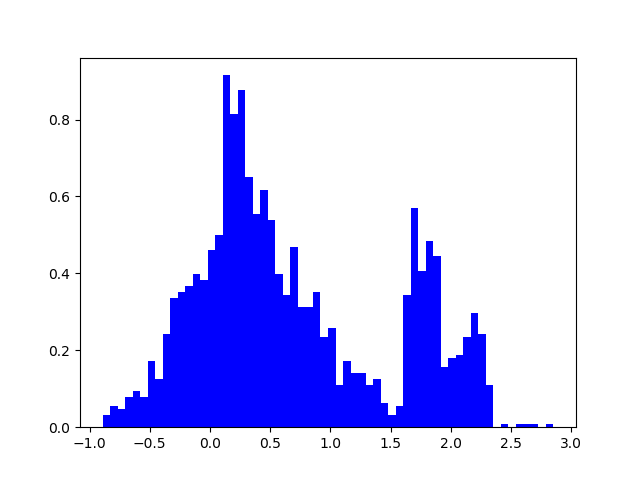}
        \caption{Source Latent (Home).}
    \end{subfigure}
    \begin{subfigure}[b]{0.49\linewidth}
        \includegraphics[width=1.0\linewidth]{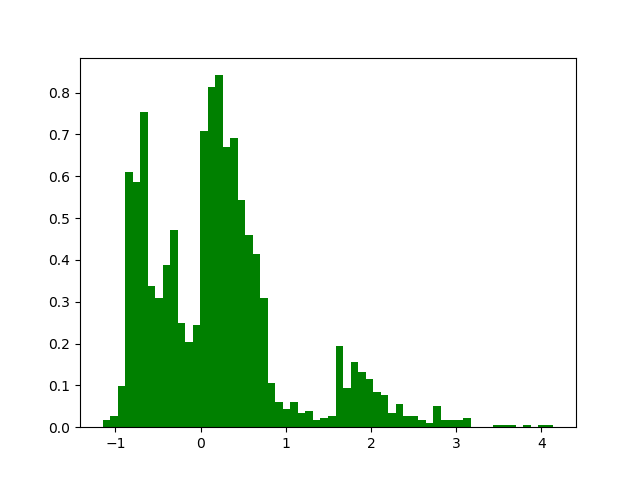}
        \caption{Target Latent (Home).}
    \end{subfigure}
    \caption{Histograms of the source latent distribution and the target latent distribution after the training of SAFN converges. \textbf{Top:} the adaptation scenario from \emph{Amazon} to \emph{DSLR} (\emph{Office-31}). \textbf{Bottom:} the adaptation scenario from \emph{Clipart} to \emph{Product} (\emph{Office-Home}).}\label{fig:shape_safn}
\end{figure}

\begin{figure}[h]
    \center
    \vspace{-4mm}
    \begin{subfigure}[b]{0.325\linewidth}
        \includegraphics[width=1.0\linewidth]{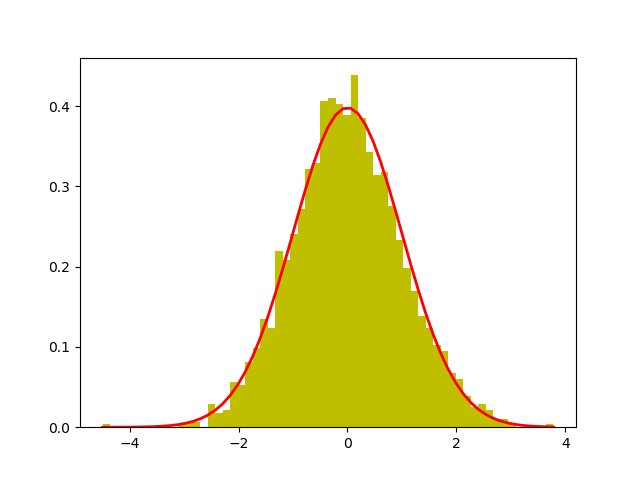}
        \caption{Gaussian Prior.}
    \end{subfigure}
    \begin{subfigure}[b]{0.325\linewidth}
        \includegraphics[width=1.0\linewidth]{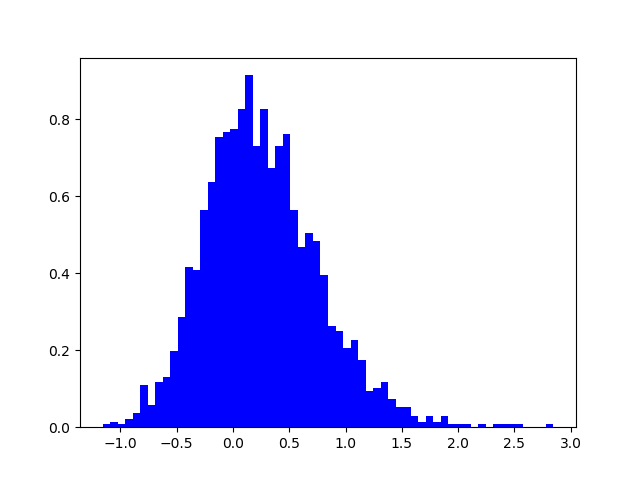}
        \caption{Source Latent.}
    \end{subfigure}
    \begin{subfigure}[b]{0.325\linewidth}
        \includegraphics[width=1.0\linewidth]{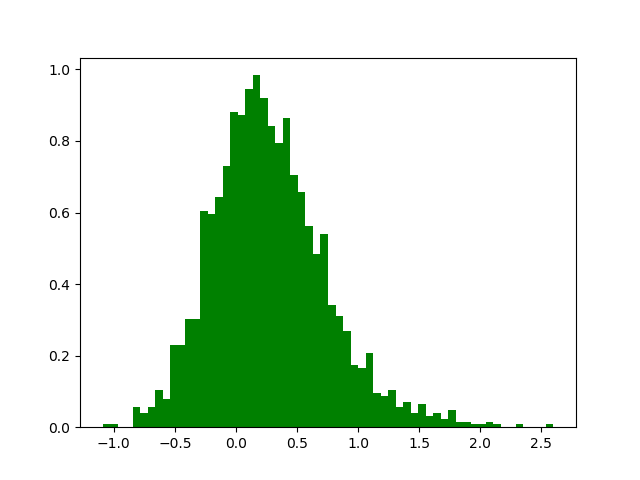}
        \caption{Target Latent.}
    \end{subfigure}
    \caption{Histograms of the source latent distribution and the target latent distribution after the training of the proposed DFA-SAFN on the adaptation scenario from \emph{Amazon} to \emph{DSLR} (\emph{Office-31}) converges. }\label{fig:shape_dfa_31}
\end{figure}

\begin{figure}[h]
    \center
    \begin{subfigure}[b]{0.325\linewidth}
        \includegraphics[width=1.0\linewidth]{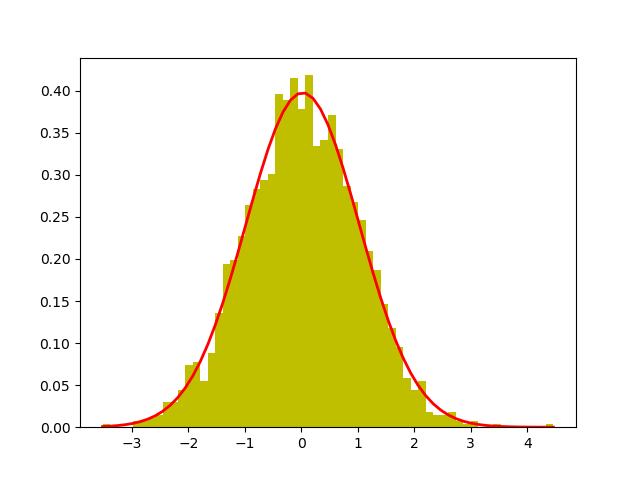}
        \caption{Gaussian Prior.}
    \end{subfigure}
    \begin{subfigure}[b]{0.325\linewidth}
        \includegraphics[width=1.0\linewidth]{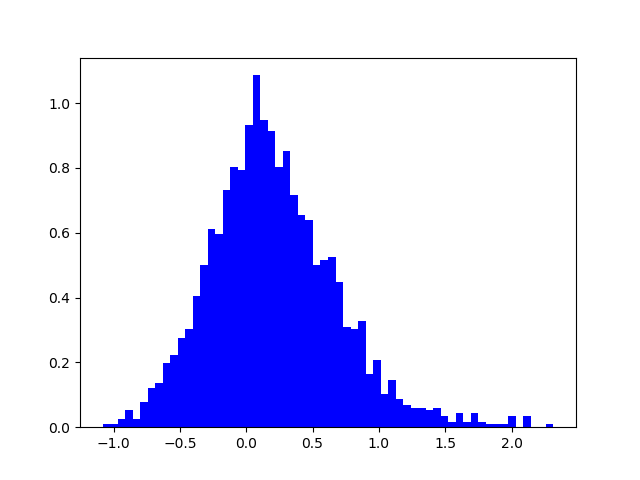}
        \caption{Source Latent.}
    \end{subfigure}
    \begin{subfigure}[b]{0.325\linewidth}
        \includegraphics[width=1.0\linewidth]{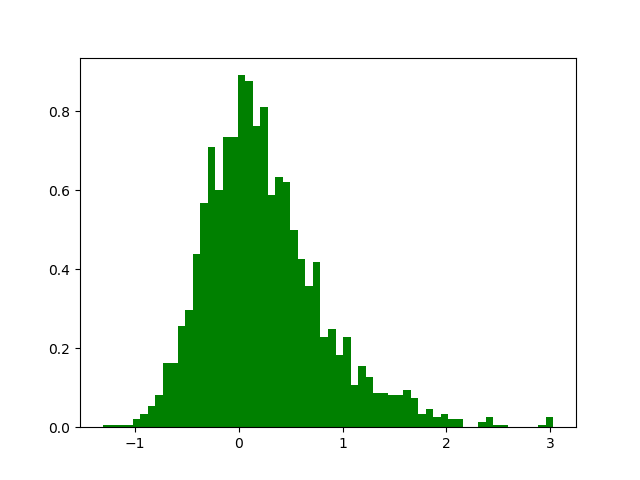}
        \caption{Target Latent.}
    \end{subfigure}
    \caption{Histograms of the source latent distribution and the target latent distribution after the training of the proposed DFA-SAFN on the adaptation scenario from \emph{Clipart} to \emph{Product} (\emph{Office-Home}) converges. }\label{fig:shape_dfa_home}
\end{figure}

In this ablation study, we validated the claim that the proposed regularization could construct the latent distributions of the two domains on a common distribution space. In our setting, the common distribution space is the space of the Gaussian prior. The best-performing models that were trained previously were used in the study. We selected a vector from the source latent distribution and one corresponding vector from the target latent distribution, and plotted their histograms for demonstration. Note that the selected vectors from the source latent distribution and the target latent distribution fall under the same category so that they should share the discriminative features. Figure~\ref{fig:shape_safn} demonstrates that the existing UDA methods (take SAFN \cite{norm} as an example) cannot effectively construct the feature spaces of the two domains on a common distribution space. This could make the classification tasks on the target samples hard to make the most use of the discriminative source features. By contrast, as shown in Figure~\ref{fig:shape_dfa_31} and Figure~\ref{fig:shape_dfa_home}, the proposed regularization can encourage the source discriminative features to be projected into the space of the Gaussian prior, and construct the target feature space on this prior distribution space. This indicates that the proposed DFA can encourage the latent distributions of the two domains to be closed to a common distribution in the feature space, i.e., the Gaussian prior, which promotes better feature alignment. Note that, the latent vectors are observed from the layer before the last ReLU activation of the encoder for better demonstration.

\subsection{Effectiveness of the Proposed Regularization} 
In this ablation study, we validated that our method could effectively align the feature spaces of the two domains. We conducted a case study on the adaptation scenario from SVHN to MNIST as its significant domain variation. We randomly selected 100 images per class from both domains and 2000 images in total. We utilized the best-performing models that were trained in the previous experiments. By measuring the distance between the feature spaces, the effectiveness of the feature alignment can be examined. We computed the average L2-distances between the feature space of SVHN and the feature space of MNIST after the adaptation with and without our model, as shown in Table~\ref{tab:latent}. As expected, the feature-space distance of DFA-MCD is much shorter than that of MCD.
\begin{table}[h]
\setlength{\tabcolsep}{3pt}
\setlength\aboverulesep{0pt}\setlength\belowrulesep{0pt}
\small
\caption{Average L2-distance between the SVHN feature space and the MNIST feature space. The numbers (\textbf{0}-\textbf{9}) denote the digit labels, and \textbf{All} indicates evaluating by all samples.}
\begin{center}
\begin{tabular}{c | c c c c c c }
\hline
\textbf{Method} & \textbf{0} & \textbf{1} & \textbf{2} &\textbf{3} & \textbf{4} & \textbf{5}\\ 
\hline
MCD & 0.1658 & 0.1433 & 0.1585 & 0.1539 & 0.1544 & 0.1598 \\
\textbf{DFA-MCD} & \textbf{0.0644} & \textbf{0.0797} & \textbf{0.0867} & \textbf{0.0879} & \textbf{0.0871} & \textbf{0.0783} \\
\hline\hline
\textbf{Method} & \textbf{6} & \textbf{7} &\textbf{8} & \textbf{9} & \textbf{All}& \\
\hline
MCD& 0.1529 & 0.1596 & 0.1472 & 0.1517 & 0.0564&\\
\textbf{DFA-MCD}& \textbf{0.0800} & \textbf{0.0829} & \textbf{0.0692} & \textbf{0.0756} & \textbf{0.0266}&\\
\hline
\end{tabular}
\end{center}
\label{tab:latent}
\vspace{-3mm}
\end{table}

\subsection{How to Effectively Align Feature Spaces}
We investigated the most effective method for the latent alignment in this ablation study. We conducted a case study on the adaptation scenario from MNIST to UPSP. To better illustrate this study, we first define some loss functions. We formulate the paired reconstruction loss of an autoencoder as:
\begin{equation}
\begin{aligned}
\mathcal{L}_{recon}(X) =  \frac{1}{M} \sum_{i=1}^{M} [\lvert\lvert D(G(\mathbf{x}^{(i)});\theta_g) - \mathbf{x}^{(i)})\lvert\lvert_{1}].
\end{aligned}
\end{equation}
We define a KL-divergence penalty to encourage $p(\mathbf{z_t})$ to be close to $p(\mathbf{z_s})$ as $\mathcal{L}_{klddir}$. To validate the effect of weight tying, we further define the learning parameters $\theta_d$ for the decoder $D$ in the case where the tying weight is not applied. We explored six different ways to align the two latent feature distributions $p(\mathbf{z_s})$ and $p(\mathbf{z_t})$: \textbf{1)} the proposed \textbf{DFA-ENT} framework; \textbf{2)} \textbf{DFA-ENT} but the encoder $G$ and the decoder $D$ do not share their weights ($\theta_d \neq \theta_g$); \textbf{3)} instead of using our DAL to align the target latent distribution with the Gaussian prior, utilizing a KL-divergence to make $p(\mathbf{z_t})$ close to the prior; \textbf{4)} the direct latent alignment via an unpaired L1-distance between the reconstructed samples from the two domains, i.e., minimizing the distance between $D(G(\mathbf{x_s}))$ and $D(G(\mathbf{x_t}))$ ($\mathcal{L}_{daldir}$); \textbf{5)} the direct latent alignment using $\mathcal{L}_{klddir}$; and \textbf{6)} further regularizing Case \textbf{5)} by two reconstruction losses $\mathcal{L}_{recon}(X_S)+\mathcal{L}_{recon}(X_T)$ ($\mathcal{L}_{recon}$) with our weight-tied encoder-decoder formulation. The results, which are shown in Table~\ref{tab:alignment}, indicate that the proposed DFA is the most effective approach to align the latent distributions of the two domains. The ablation study validates that all of the Gaussian-guided alignment, unpaired L1-distance and weight tying are of necessity for the proposed model. 

\begin{table}[h]
\setlength{\tabcolsep}{3pt}
\setlength\aboverulesep{0pt}\setlength\belowrulesep{0pt}
\small
\caption{Accuracy(\%) of different latent-alignment methods on the adaptation scenario from MNIST to USPS. Note that all methods utilize $\mathcal{L}_{ent}$ and $\mathcal{L}_{cls}$ for classification.}

\begin{center}
\begin{tabular}{c | c c c}
\hline
& $\mathcal{L}_{kld} + \mathcal{L}_{dal}$(Ours) & $\mathcal{L}_{daldir}$ & $\mathcal{L}_{kld}$ \\ 
\hline
\textbf{Accuracy} & \textbf{97.3} & 93.1 & 87.9 \\
\hline
\hline
& $\mathcal{L}_{kld} + \mathcal{L}_{dal}, \theta_d \neq \theta_g$ & $\mathcal{L}_{klddir}$ & $\mathcal{L}_{klddir}+\mathcal{L}_{recon}$\\ 
\hline
\textbf{Accuracy} & 95.8 & 89.2 & 83.6 \\
\hline
\end{tabular}
\end{center}
\label{tab:alignment}
\vspace{-3mm}
\end{table}

\subsection{Parameter Sensitivity}
\begin{figure}[h]
    \center
    \begin{subfigure}[b]{0.45\linewidth}
        \includegraphics[width=1.0\linewidth]{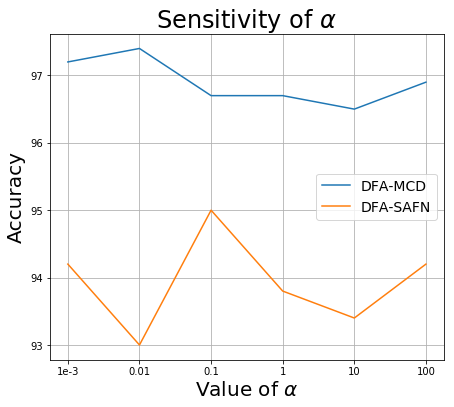}
        \caption{$\alpha$}
    \end{subfigure}
    \begin{subfigure}[b]{0.45\linewidth}
        \includegraphics[width=1.0\linewidth]{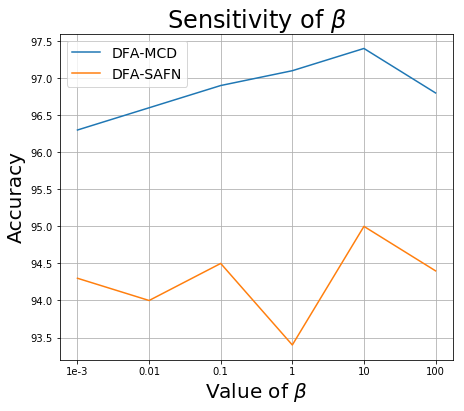}
        \caption{$\beta$}
    \end{subfigure}
    \caption{Sensitivity analysis of the hyper-parameters $\alpha$ and $\beta$ for DFA-MCD and DFA-SAFN (\textbf{orange} lines indicate DFA-SAFN; \textbf{blue} lines indicate DFA-MCD). $\alpha$ was set to 0.1 when evaluating $\beta$. $\beta$ was set to 10 when evaluating $\alpha$.}\label{fig:ablation}
\end{figure}

To quantify the impact of our \emph{discriminative feature alignment} (DFA) on the UDA frameworks, we investigated the sensitivity of our hyper-parameters, i.e., $\alpha$ and $\beta$, in DFA-MCD and DFA-SAFN. We selected adaptation scenarios from MNIST to USPS and from Amazon to DSLR for demonstration. The results are shown in Figure~\ref{fig:ablation}(a)(b). For each case study, $\alpha$ and $\beta$ were varied from 0.001 to 100. As shown in both figures, DFA can stably improve the performance of adversarial and non-adversarial UDA frameworks with different values of $\alpha$ and $\beta$.

\subsection{Computational Complexity Analysis}
\begin{figure}[h]
    \center
    \begin{subfigure}[b]{0.67\linewidth}
        \includegraphics[width=1.0\linewidth]{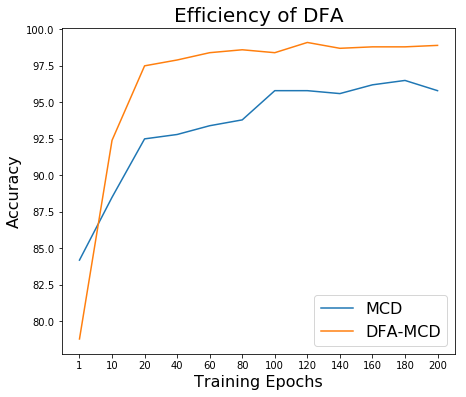}
    \end{subfigure}
    \caption{Relationship between the training epoch and the accuracy (\textbf{orange} line indicates the proposed DFA-MCD; \textbf{blue} line indicates MCD).}\label{fig:efficency}
    \vspace{-3mm}
\end{figure}

We investigated the computational efficiency of our model as it could be combined with other UDA frameworks. We conducted a case study on the adaptation scenario from SVHN to MNIST. Although the time spent on training one epoch for DFA-MCD is $1.21$ times MCD (NVIDIA GeForce RTX 2070), DFA-MCD requires fewer epochs to converge, as shown in Figure~\ref{fig:efficency}. Therefore, we can say that our model can efficiently improve the performance of various UDA frameworks. 

\section{Conclusion} 
In this paper, we introduced a novel model for UDA to better align the source and the target features, which could improve the adaptation performance of the UDA framework. We proposed an indirect latent alignment process to encourage the features of the two domains to be constructed on a common feature space, i.e., the space of the Gaussian prior. To better align two distributions, we also proposed a novel unpaired L1-distance in the decoder space, and empirically confirmed that it served as a distribution alignment mechanism. Our frameworks outperformed state-of-the-arts in most experiments. The results of the extensive experiments have validated the importance and the versatility of our research.

\bibliographystyle{cas-model2-names}

\bibliography{cas-refs}

\end{document}